\documentclass[10pt,twocolumn,letterpaper]{article}

 \usepackage[pagenumbers]{cvpr} % To force page numbers, e.g. for an arXiv version

\definecolor{cvprblue}{rgb}{0.21,0.49,0.74}
\usepackage[pagebackref,breaklinks,colorlinks,allcolors=cvprblue]{hyperref}
\usepackage{amsthm}
\usepackage{multirow}
\usepackage{subcaption}
\usepackage{algorithm}
\usepackage{algorithmic}
\usepackage[table]{xcolor}
\definecolor{lightblue}{RGB}{220,235,250}

\newcommand{\supgain}[1]{%
    {\raisebox{0.4ex}{\scriptsize\textcolor{green!50!black}{$\uparrow$#1}}}
}

\newtheorem{theorem}{Theorem}
\newtheorem{assumption}{Assumption}

\def\confName{CVPR}
\def\confYear{2027}

\title{Risk-Controlled Test-Time Adaptation for Vision-Language-Action Models}

\author{
  Zixuan Liu$^{1}$ \quad
  Hashem Haghbayan$^{2}$ \quad
  Zizhan Zheng$^{1}$ \quad
  Siavash Khajavi$^{3}$ \\
  $^{1}$Department of Computer Science, Tulane University \\
  $^{2}$Department of Computing, University of Turku \\
  $^{3}$Department of Industrial Engineering and Management, Aalto University \\
  {\tt\small \{zliu41, zzheng3\}@tulane.edu} \\
  {\tt\small hashem.haghbayan@utu.fi} \\
  {\tt\small siavash.khajavi@aalto.fi}
}

\begin{document}
\maketitle
\begin{abstract}

Retrieval-augmented vision-language-action (VLA) models enable efficient test-time adaptation by leveraging expert demonstrations as additional context without updating model parameters. However, even when retrieved demonstrations are relevant to the current task, the VLA may fail to effectively utilize the retrieved information, ultimately leading to task failure. To address this challenge, we propose \textbf{R2-VLA}, a risk-controlled retrieval-augmented VLA framework that enables reliable test-time adaptation by satisfying a user-specified task success rate requirement with minimal expert assistance. Specifically, R2-VLA learns an uncertainty predictor that estimates the uncertainty of the VLA's predicted actions given the retrieved demonstrations, guiding when to act autonomously or request expert assistance. We further introduce a task-level risk-control procedure that calibrates the uncertainty threshold to satisfy a user-specified task-failure risk requirement. Moreover, R2-VLA continually incorporates expert corrections into its retrieval memory to improve future adaptation while preserving the prescribed task-level risk guarantee. Experiments across three VLA backbones and four LIBERO benchmark suites demonstrate that R2-VLA improves average task success by \textbf{9.2--11.2} percentage points, achieving the desired target with fewer expert requests. These results highlight the effectiveness of connecting the usefulness of retrieved demonstrations with task-level risk control to enable successful robotic manipulation.

\end{abstract}    
\section{Introduction}
\label{sec:intro}

\begin{figure*}[t]
    \centering
    \includegraphics[width=\linewidth]{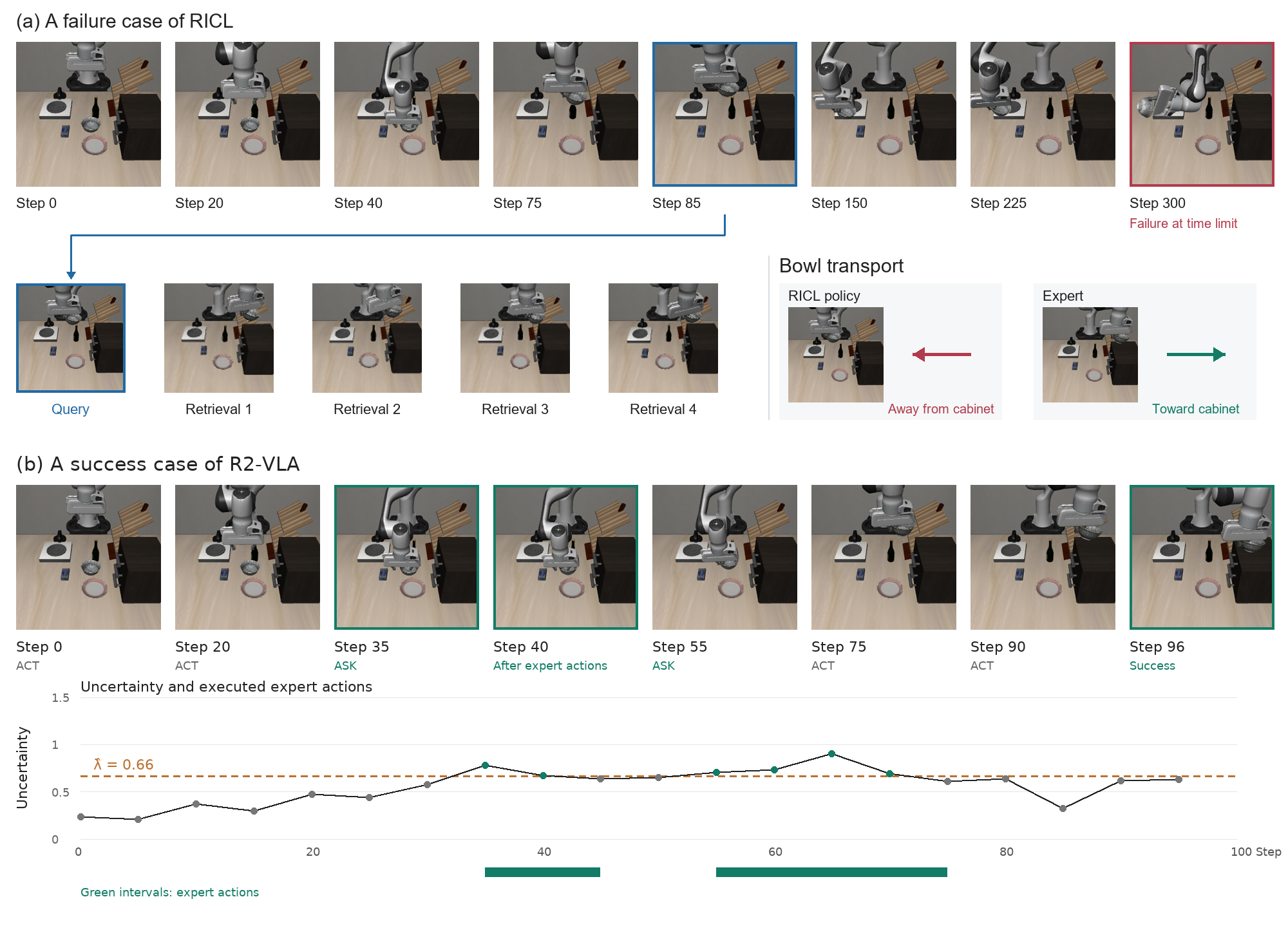}
    \caption{
    Qualitative comparison of RICL and R2-VLA on the LIBERO-\textsc{Goal} task \emph{``Put the bowl on top of the cabinet''} using $\pi_0$-FAST. (a) RICL retrieves behaviorally relevant demonstrations from the same task, yet still produces an incorrect action at step 85: instead of moving the bowl toward the cabinet, the policy moves it away from the target. This error leads to continued execution in the wrong direction and ultimately task failure. (b) R2-VLA detects such uncertain actions using its learned uncertainty predictor and requests expert assistance when the predicted uncertainty exceeds the threshold $\hat \lambda$, enabling successful task completion.
    }
    \label{fig:qualitative}
    \vspace{-2ex}
\end{figure*}

Recent advances in vision-language-action (VLA) models~\cite{black2024pi_0,kim2024openvla,pertsch2025fast,team2024octo,bousmalis2023robocat,brohan2023rt,vuong2023open,team2025gemini} have enabled a single policy to perform a broad range of robotic manipulation tasks, highlighting their potential as general-purpose robotic control policies. However, even strong pre-trained VLAs often require substantial fine-tuning or retraining to maintain reliable performance under distribution shifts, such as new environments, unseen objects, or novel task instructions~\cite{zhang2026retrieval}, limiting their ability to rapidly adapt at deployment time. In contrast, Large Language Models (LLMs) can adapt to new tasks at inference time through in-context learning (ICL): given only a small number of input-output demonstrations as context, an LLM can adjust its behavior without updating its model parameters~\cite{brown2020language,dong2024survey,mao2025survey}. Inspired by this paradigm, recent work such as RICL~\cite{sridhar2025ricl} introduces in-context adaptation to pre-trained VLAs. Specifically, RICL maintains a compact retrieval buffer of expert demonstrations and, at each decision step, retrieves demonstrations relevant to the current observation as additional context for the VLA policy, enabling effective adaptation to previously unseen deployment conditions without parameter updates.

The effectiveness of retrieval-based in-context adaptation depends critically on which demonstrations are retrieved, since these examples determine what task-specific information is provided to the VLA at deployment. Prior work~\cite{sridhar2025ricl,zhang2026retrieval}, including RICL, relies on visual similarity between the current observation and observations in the retrieval buffer to select relevant demonstrations. Recent approaches have explored improved retrieval mechanisms~\cite{jang2026ra,yang2026ici,liu2026robota}, such as incorporating task instructions and learning embeddings that better capture behavioral relevance. However, improving retrieval quality does not necessarily guarantee successful task execution~\cite{liu2026does}. Even when the retrieved demonstrations are relevant to the current task, the VLA may still fail to translate the retrieved information into appropriate actions for its current state. As illustrated in Fig.~\ref{fig:qualitative}(a), RICL retrieves behaviorally relevant demonstrations but nevertheless moves the bowl in the wrong direction, ultimately leading to task failure. This suggests the usefulness of retrieved demonstrations depends not solely on their relevance to the current task, but ultimately on whether they enable the VLA to successfully complete it.

In this work, we propose \textbf{R2-VLA}, a risk-controlled retrieval-augmented VLA framework that enables reliable test-time adaptation by satisfying a user-specified task success rate. As discussed above, retrieving relevant demonstrations alone is insufficient for successful task execution, as the VLA may still fail to effectively utilize the retrieved information. Therefore, R2-VLA enables the VLA to determine when it can act autonomously and when additional assistance is necessary to successfully complete the task. To this end, we augment the VLA with a learned uncertainty predictor that estimates how uncertain the VLA is about its predicted actions, given the current observation, retrieved demonstrations, and its own prior knowledge. When the predicted uncertainty is below a specified threshold, the VLA executes its own action autonomously (\textsc{Act}). Otherwise, it requests expert assistance (\textsc{Ask}) to avoid potentially incorrect actions, as illustrated in Fig.~\ref{fig:qualitative} (b). However, determining an appropriate uncertainty threshold is nontrivial, as the predicted uncertainty does not directly correspond to task-level success. We therefore introduce a risk-control procedure that calibrates the threshold to satisfy a user-specified task-failure risk requirement. In this way, R2-VLA directly connects the usefulness of retrieved demonstrations to their ability to support successful task completion.

To summarize, our contributions are as follows: 1. We propose \textbf{R2-VLA}, a risk-controlled retrieval-augmented VLA framework that learns to estimate the uncertainty of its predicted actions given retrieved demonstrations and determines when to \textsc{Act} or \textsc{Ask} to satisfy a user-specified task-level failure risk requirement. 2. We theoretically establish a connection between uncertainty prediction accuracy and task-level failure risk (Theorem~\ref{thm:risk-bound}). We further extend R2-VLA to continual adaptation by incorporating expert corrections into the retrieval buffer and theoretically establish that the prescribed task-level risk can be preserved throughout adaptation (Theorem~\ref{thm:continual-risk-main}). 3. We conduct extensive experiments across three VLA backbones and four LIBERO benchmark suites, demonstrating consistent improvements of \(9.2\)--\(11.2\) percentage points in average task success over RICL, while meeting the desired target in most settings with minimal expert assistance.

\section{Preliminary}
\label{sec:preliminary}

\noindent\textbf{Problem Setup.}
In this paper, we consider a general robotic manipulation setting~\cite{zhang2026retrieval,sridhar2025ricl}, where a pre-trained vision-language-action (VLA) policy \(\pi_\theta\), parameterized by \(\theta\), is required to perform novel tasks in unseen environments without any test-time parameter updates. Each task is specified by a natural language instruction \(l^{\mathrm{text}}\). At timestep \(t\), the agent observes multi-view images \(I_t=(I_t^{\mathrm{top}},I_t^{\mathrm{side}},I_t^{\mathrm{wrist}})\), corresponding to the top, side, and wrist camera views, together with a proprioceptive state \(s_t\). A standard VLA policy predicts an action according to \(\hat a_t \sim \pi_\theta(\cdot \mid x_t)\), where \(x_t=(I_t,s_t,l^{\mathrm{text}})\) denotes the current input.

\noindent\textbf{Re-training for In-Context Learning (RICL).}
To improve test-time adaptability without updating model parameters,~\cite{sridhar2025ricl} introduce Re-training for In-Context Learning (RICL), which conditions the VLA on retrieved expert demonstrations provided as in-context examples. Let \(\mathcal D_{\mathrm{ret}}=\{m_i\}_{i=1}^{N_{\mathrm{ret}}}\) denote a retrieval buffer containing \(N_{\mathrm{ret}}\) expert demonstration frames, where each entry is \(m_i=(I_i,s_i,l_i^{\mathrm{text}},a_i)\). Here, \(I_i=(I_i^{\mathrm{top}},I_i^{\mathrm{side}},I_i^{\mathrm{wrist}})\) denotes the multi-view observations at the demonstration state, \(s_i\) is the corresponding proprioceptive state, \(l_i^{\mathrm{text}}\) is the task instruction, and \(a_i\) is the expert action or action chunk, i.e., an array of actions over multiple timesteps. These demonstrations are not used to update the VLA parameters at deployment. Instead, they provide external context for adapting the policy to the current task. At each timestep \(t\) during deployment, given the current input \(x_t=(I_t,s_t,l^{\mathrm{text}})\), RICL retrieves a small set of relevant entries from \(\mathcal D_{\mathrm{ret}}\). Let \(f_{\mathrm{DINO}}\) denote a frozen DINOv2 image encoder~\cite{oquab2023dinov2}. RICL uses the top-view image \(I_t^{\mathrm{top}}\) as the retrieval query and compares it with the top-view image of each demonstration entry. Specifically, for each \(m_i\in\mathcal D_{\mathrm{ret}}\), it computes the distance \(d_t(i)=\|f_{\mathrm{DINO}}(I_t^{\mathrm{top}})-f_{\mathrm{DINO}}(I_i^{\mathrm{top}})\|_2\). The \(M\) nearest entries are then selected as \((j_1,\ldots,j_M)=\operatorname{TopM}_{i\in\{1,\ldots,N_{\mathrm{ret}}\}}^{\mathrm{smallest}} d_t(i)\), with \(d_t(j_1)\le d_t(j_2)\le \cdots \le d_t(j_M)\). The resulting retrieved context is \(\mathcal M_t=(m_{j_1},\ldots,m_{j_M})\), where entries are ordered from the nearest to the farthest neighbor. RICL then conditions on both the retrieved context and the current input to predict the next action, \(\hat a_t \sim \pi_\theta(\cdot\mid \mathcal M_t,x_t)\). 

\noindent\textbf{Risk Control.}
Risk control provides a general framework for calibrating a decision-making system so that a user-specified notion of risk remains below a desired level~\cite{bates2021distribution,angelopoulos2025learn,angelopoulos2024conformal}. In general, risk-control methods use a held-out calibration dataset to determine whether a decision rule satisfies a prescribed population-level risk requirement. Formally, consider a family of decision rules \(\{\pi^{\lambda}:\lambda\in\Lambda\}\) indexed by a calibration parameter \(\lambda\), and let \(L(\pi^{\lambda};Z)\in[0,1]\) denote the loss incurred when applying \(\pi^{\lambda}\) to a random example \(Z\). The corresponding population risk is \(R(\lambda)=\mathbb E_{Z\sim\mathcal P}\left[L(\pi^{\lambda};Z)\right]\), where \(\mathcal P\) denotes the deployment distribution. Given a user-specified target risk \(\epsilon\), the goal is to use a held-out calibration set \(\mathcal D_{\mathrm{cal}}\) to select a parameter \(\hat\lambda\) such that \(R(\hat\lambda)\leq\epsilon\).

We provide a more detailed discussion of related work on retrieval-augmented VLA and risk control in Appendix~\ref{app:related-work}.
\section{Risk-Controlled Retrieval-Augmented VLA}

Even behaviorally relevant demonstrations do not necessarily enable a retrieval-augmented VLA to complete a task successfully. We therefore propose \textbf{R2-VLA}, which determines when the VLA can act autonomously and when it should request expert assistance, with the goal of satisfying a user-specified task success requirement while minimizing the frequency of expert assistance. We first formulate the R2-VLA policy (Section~\ref{sec:framework}), then describe how the uncertainty predictor is learned (Section~\ref{sec:uncertain}) and how its threshold is calibrated using task outcomes (Section~\ref{sec:risk-control}). Finally, we introduce continual deployment adaptation (Section~\ref{sec:algorithm}) and present the theoretical results (Section~\ref{sec:theory}).

\subsection{R2-VLA Framework}
\label{sec:framework}

R2-VLA adopts an uncertainty-guided \textsc{Act}/\textsc{Ask} decision framework built upon a retrieval-augmented VLA policy $\pi_\theta$ with a fixed retrieval buffer $\mathcal D_{\mathrm{ret}}$. At each decision step $t$, the current input $x_t$ and retrieved context $\mathcal M_t$ are provided to the VLA, which predicts an action $\hat a_t$. We augment the VLA with a lightweight uncertainty predictor $U_\psi$ that outputs a scalar uncertainty score $\hat u_t=U_\psi(\mathcal M_t,x_t)>0$ from its internal representations, reflecting the VLA's own assessment of the uncertainty associated with its predicted action. Thus, the augmented model produces $(\hat a_t, \hat u_t)=\pi_{\theta,\psi}(\cdot\mid\mathcal M_t,x_t)$. Given a threshold $\lambda$ that specifies the maximum tolerable predicted action uncertainty, R2-VLA follows the decision rule:
\[
\pi_{\theta,\psi}^{\lambda}(\mathcal M_t,x_t)
=
\begin{cases}
\hat a_t, & \hat u_t\leq\lambda \quad (\text{\textsc{Act}}),\\[1mm]
a_t^{\mathrm E}, & \hat u_t>\lambda \quad (\text{\textsc{Ask}}),
\end{cases}
\]
where $a_t^{\mathrm E}$ denotes the expert action provided only when \textsc{Ask} is triggered. A lower threshold generally makes the policy more cautious, whereas a higher threshold permits autonomous execution at more decision points.

The objective of R2-VLA is to satisfy a user-specified task success rate requirement while minimizing the frequency of expert assistance. Formally, let $Z=(\xi,l^{\mathrm{text}})$ denote a task instance, where $\xi$ represents the initial environment configuration, including the initial states of the robot and objects, and $l^{\mathrm{text}}$ denotes the task instruction. Let $\mathcal P$ denote the deployment distribution over task instances. Executing $\pi_{\theta,\psi}^{\lambda}$ from $Z$ produces a trajectory $\tau^\lambda(Z)$ of $T$ decision steps. We define the task-level failure loss as \(L(\lambda;Z)=\mathbb I\{\tau^\lambda(Z)\text{ fails}\}\), and the corresponding population risk as \(R(\lambda)=\mathbb E_{Z\sim\mathcal P}[L(\lambda;Z)].\) For the same trajectory, define its ASK rate as \(H(\lambda;Z)=\frac{1}{T}\sum_{t=1}^{T}\mathbb I\{\hat u_t>\lambda\},\) and its average as \(H(\lambda)=\mathbb E_{Z\sim\mathcal P}[H(\lambda;Z)].\) Given a user-specified maximum task-failure risk $\epsilon$, equivalently a target success rate of $1-\epsilon$, our objective is to select a threshold from the candidate set $\Lambda$ that minimizes the frequency of expert assistance while satisfying the prescribed task-level risk requirement. Formally, we solve:
\begin{equation}
\label{eq:r2vla-objective}
\min_{\lambda\in\Lambda}\quad  H(\lambda) \qquad
\text{subject to}\quad  R(\lambda)\leq\epsilon.
\end{equation}
Solving Eq.~\eqref{eq:r2vla-objective} is challenging for two reasons. First, the deployment distribution $\mathcal P$ is unknown, making it difficult to directly evaluate the population risk $R(\lambda)$ and ASK rate $H(\lambda)$. Second, changing the threshold can alter subsequent trajectories, so $H(\lambda)$ is not monotonic in $\lambda$, requiring us to evaluate each candidate threshold in $\Lambda$. To address these challenges, we next describe how R2-VLA learns an uncertainty predictor to guide \textsc{Act}/\textsc{Ask} decisions and uses held-out task rollouts to select a statistically certified threshold that minimizes the empirical ASK rate among candidates.

\subsection{Learning the Uncertainty Predictor}
\label{sec:uncertain}

As discussed above, the uncertainty predictor provides a per-step signal to guide \textsc{Act}/\textsc{Ask} decisions by assessing the VLA's uncertainty about its predicted action from its internal representations, rather than relying on external retrieval quality metrics, such as the DINO distance used by RICL. We characterize action uncertainty through the discrepancy between the VLA's predicted action and the corresponding expert action, where a large discrepancy indicates that the VLA's predicted action deviates from the desired expert behavior, potentially leading to task failure. We therefore train the uncertainty predictor to estimate this discrepancy, as detailed below.

We first train the original retrieval-augmented RICL VLA $\pi_\theta$ using a predefined retrieval buffer $\mathcal D_{\mathrm{ret}}$ and the corresponding training data, as detailed in Appendix~\ref{app:vla-backbone}. After the VLA has been trained, we freeze $\theta$ and optimize only $U_\psi$. This separation ensures that the target used to train the uncertainty head remains fixed. If \(\pi_\theta\) were updated simultaneously, its predicted action \(\hat a_t\) would continuously change during training and cause the moving target problem~\cite{fahlman1989cascade}. At training step $t$, the frozen VLA predicts $\hat a_t$ given $(\mathcal M_t,x_t)$. Let $a_t^{\mathrm E}$ denote the corresponding expert action, which serves as the ground-truth reference for uncertainty learning. We define the action uncertainty $u_t$ as the normalized discrepancy between the VLA's predicted action $\hat a_t$ and the expert action $a_t^{\mathrm E}$:
\[
u_t=d_A(\hat a_t,a_t^{\mathrm E})
=
\sqrt{\frac{1}{HD}\sum_{h=1}^{H}\sum_{d=1}^{D}
\left(\frac{\hat a_{t,h,d}-a_{t,h,d}^{\mathrm E}}{\sigma_d}\right)^2}.
\]
where $D$ is the action dimension, $H$ is the number of actions in an action chunk, and $\sigma_d$ is the standard deviation of action dimension $d$, estimated once from the training data. This normalization accounts for the fact that different action dimensions, such as translation, rotation, and gripper control, may operate on substantially different numerical scales. We train the uncertainty predictor to approximate the action discrepancy, i.e., \(\hat u_t \approx u_t\), by minimizing the following uncertainty regression loss
\[
\mathcal L_U
=
\frac{1}{\mathbb E[T]}
\mathbb E\left[\sum_{t=1}^{T}
\mathrm{SmoothL1}\bigl(\log(\hat u_t+\eta),\log(u_t+\eta)\bigr)
\right],
\]
where \(\eta=10^{-4}>0\) is a numerical stabilization constant.

A key challenge is ensuring that the uncertainty head learns failure modes specifically induced by retrieval. If it is trained only with ordinary retrieval contexts from \(\mathcal D_{\mathrm{ret}}\), most retrieved demonstrations may be helpful, providing relatively few examples in which retrieval actively degrades the resulting action. We therefore augment uncertainty-head training with deliberately degraded retrieval contexts, including wrong-task retrieval, visually similar retrieval associated with a different goal, and random or uninformative retrieval, as detailed in Appendix~\ref{app:uncertainty-head} Table~\ref{tab:composition-corrupt}. Crucially, these corrupted contexts are used to teach the uncertainty head \emph{action uncertainty}, rather than retrieval abnormality. A poor or unusual retrieval should not necessarily result in high uncertainty if the VLA can still produce the correct action from its pretrained knowledge. In this case, \(u_t\approx 0\), and the desired uncertainty remains small. Conversely, a retrieved context may appear behavior relevant while steering the VLA toward an incorrect behavior, in which case \(u_t\gg 0\) and the uncertainty should be correspondingly large. We provide an ablation study on the effect of incorporating corrupted retrieval contexts during uncertainty-head training in Appendix~\ref{app:ablation}, Table~\ref{tab:ablation-corruption-all}.

\subsection{Risk-Controlled Calibration}
\label{sec:risk-control}

After training the uncertainty predictor, R2-VLA produces both an action prediction and a predicted uncertainty score. During deployment, all model parameters remain frozen, and the policy executes its predicted action when $\hat u_t \leq \lambda$ and requests expert assistance otherwise, as defined in Sec.~\ref{sec:framework}. To solve the threshold selection problem in Eq.~\eqref{eq:r2vla-objective}, we adopt the risk-control framework introduced in Sec.~\ref{sec:preliminary} to identify thresholds that satisfy the prescribed task-failure risk requirement with high probability and select the one with the lowest empirical ASK rate among the certified candidates. Specifically, since the deployment distribution $\mathcal P$ is unknown, we construct a held-out calibration set $\mathcal D_{\mathrm{cal}}=\{Z_i\}_{i=1}^{N_{\mathrm{cal}}}$ to empirically estimate the population risk $R(\lambda)$ and ASK rate $H(\lambda)$. The calibration set consists of independent task instances sampled from $\mathcal P$ and is disjoint from the data used for retrieval and uncertainty learning. Each task instance $Z_i=(\xi_i,l_i^{\mathrm{text}})$ specifies an initial environment configuration and task instruction. We consider a sufficiently large, predefined candidate threshold set $\Lambda=\{\lambda_1,\ldots,\lambda_K\}$, with details on its construction provided in Appendix~\ref{app:benchmark-eval}. Since $H(\lambda)$ is not monotonic in $\lambda$, we estimate both $R(\lambda)$ and $H(\lambda)$ for each candidate threshold $\lambda\in\Lambda$ to identify the best $\lambda$ for solving Eq.~\eqref{eq:r2vla-objective}.

The complete procedure is summarized in Algorithm~\ref{alg:risk-control}. For each $\lambda\in\Lambda$ and calibration instance $Z_i$, we reset the environment to $\xi_i$ and execute the R2-VLA policy $\pi_{\theta,\psi}^{\lambda}$ until termination, producing a trajectory $\tau_i^\lambda$. We record the episode-level failure loss as $L_i(\lambda)=\mathbb I\{\tau_i^\lambda\text{ fails}\}$ and compute the total number of failed calibration episodes as $F_\lambda=\sum_{i=1}^{N_{\mathrm{cal}}}L_i(\lambda)$ (Algorithm~\ref{alg:risk-control}, Lines 4--6). We then determine whether the observed failure count \(F_\lambda\) is sufficiently small to reject the hypothesis that the task-failure risk exceeds the prescribed target \(\epsilon\). Following~\cite{angelopoulos2025learn}, we formulate this as the hypothesis test \(\mathcal H_{0,\lambda}:R(\lambda)>\epsilon\). Thus, certifying \(\lambda\) amounts to finding sufficient evidence against this hypothesis. Because each episode-level loss is binary, if the true failure probability were exactly equal to the target \(\epsilon\), then the number of failures among \(N_{\mathrm{cal}}\) independent calibration episodes would follow \(B\sim\operatorname{Binomial}(N_{\mathrm{cal}},\epsilon)\). We therefore compute \(p_\lambda=\Pr(B\leq F_\lambda)\) (Algorithm~\ref{alg:risk-control}, Line 9). This quantity measures how likely it would be to observe \(F_\lambda\) or fewer failures if the true failure rate were at \(\epsilon\). If \(p_\lambda\) is small, the observed failure count is unusually low even under this boundary case, providing evidence against the hypothesis \(\mathcal H_{0,\lambda}\).  

Since we evaluate multiple candidate thresholds, we must also control the probability of falsely certifying any threshold whose task-failure risk exceeds \(\epsilon\). Using Bonferroni correction~\cite{angelopoulos2025learn} with confidence parameter \(\delta\), we construct the  certified set $\widehat{\Lambda}_{\mathrm{safe}}\subseteq\Lambda$, which contains all candidate thresholds that pass the statistical risk-control test: \(\widehat{\Lambda}_{\mathrm{safe}}=\left\{\lambda\in\Lambda:p_\lambda\leq\frac{\delta}{K}\right\}\) (Algorithm~\ref{alg:risk-control}, Line 12). The correction allocates the total error probability \(\delta\) across the \(K\) hypothesis tests. Consequently, with probability at least \(1-\delta\), every threshold in \(\widehat{\Lambda}_{\mathrm{safe}}\) satisfies \(R(\lambda)\leq\epsilon\). A detailed justification is provided in Appendix~\ref{app:proof-theorem1}, Theorem~\ref{thm:single-risk}. For calibration episode \(Z_i\), let $T_i$ denote its number of decision steps and $\hat u_{i,t}$ the predicted uncertainty at step $t$. Its observed ASK rate is $H_i(\lambda)=\frac{1}{T_i}\sum_{t=1}^{T_i}\mathbb I\{\hat u_{i,t}>\lambda\}$ (Algorithm~\ref{alg:risk-control}, Line 7), and the average calibration ASK rate is $\widehat H(\lambda)=\frac{1}{N_{\mathrm{cal}}}\sum_{i=1}^{N_{\mathrm{cal}}}H_i(\lambda)$ (Algorithm~\ref{alg:risk-control}, Line 10). We then choose \(\hat\lambda=\arg\min_{\lambda\in\widehat{\Lambda}_{\mathrm{safe}}}\widehat H(\lambda)\) (Algorithm~\ref{alg:risk-control}, Line 13). Thus, $\hat\lambda$ provides an approximate solution to Eq.~\eqref{eq:r2vla-objective}, minimizing the empirical ASK rate among thresholds that satisfy the statistical risk-control criterion.

\begin{algorithm}[t]
\caption{Risk-Controlled Calibration of the \textsc{Act}/\textsc{Ask} Threshold}
\label{alg:risk-control}
\begin{algorithmic}[1]
\REQUIRE Fixed retrieval buffer $\mathcal D_{\mathrm{ret}}$, calibration set $\mathcal D_{\mathrm{cal}}$, candidate thresholds $\Lambda=\{\lambda_1,\ldots,\lambda_K\}$, R2-VLA policy $\{\pi_{\theta,\psi}^{\lambda}:\lambda\in\Lambda\}$, target failure rate $\epsilon$, confidence level $\delta$
\ENSURE Calibrated threshold $\hat\lambda$

\FOR{$\lambda\in\Lambda$}
    \STATE $F_\lambda \leftarrow 0$, $\mathcal H_\lambda \leftarrow 0$
    \FOR{$Z_i=(\xi_i,l_i^{\mathrm{text}})\in\mathcal D_{\mathrm{cal}}$}
        \STATE Reset environment to $\xi_i$, execute $\pi_{\theta,\psi}^{\lambda}$ using $\mathcal D_{\mathrm{ret}}$ and record trajectory $\tau_i^\lambda$ of length $T_i$
        \STATE $L_i(\lambda)\leftarrow
        \mathbb I\{\tau_i^\lambda\text{ fails}\}$
        \STATE $F_\lambda\leftarrow F_\lambda+L_i(\lambda)$
        \STATE Compute $H_i(\lambda)=
        \frac{1}{T_i}
        \sum_{t=1}^{T_i}
        \mathbb I\{\hat u_{i,t}>\lambda\}$, where $u_{i,t}$ is the uncertainty at step $t$
    \ENDFOR
    \STATE $p_\lambda\leftarrow
    \Pr\!\left[
    B\leq F_\lambda
    \right]$,
    $B\sim\operatorname{Binomial}(N_{\mathrm{cal}},\epsilon)$
    \STATE $\widehat H(\lambda)\leftarrow
    \frac{1}{N_{\mathrm{cal}}}
    \sum_{i=1}^{N_{\mathrm{cal}}}H_i(\lambda)$
\ENDFOR

\STATE $\widehat{\Lambda}_{\mathrm{safe}}
\leftarrow
\left\{
\lambda\in\Lambda:
p_\lambda\leq\frac{\delta}{K}
\right\}$

\STATE $\hat\lambda
\leftarrow
\arg\min_{\lambda\in\widehat{\Lambda}_{\mathrm{safe}}}
\widehat H(\lambda)$

\RETURN $\hat\lambda$
\end{algorithmic}
\end{algorithm}

\subsection{Continual Deployment Adaptation}
\label{sec:algorithm}

The risk-control procedure in Sec.~\ref{sec:risk-control} calibrates the threshold \(\hat\lambda\) for a fixed retrieval buffer \(\mathcal D_{\mathrm{ret}}\). During deployment, however, every \textsc{Ask} decision provides new expert experience at states where the current retrieval memory is insufficient. We therefore use these expert corrections to continually expand the retrieval buffer and improve future adaptation as summarized in Algorithm~\ref{alg:continual}. Formally, we organize deployment into adaptation rounds indexed by \(q=0,1,\ldots, Q\). In practice, a new adaptation round can be triggered after a fixed number of deployment rollouts. At the beginning of round \(q\), R2-VLA uses a fixed retrieval buffer \(\mathcal D_{\mathrm{ret}}^{(q)}\) together with its calibrated threshold \(\hat\lambda^{(q)}\) (Algorithm~\ref{alg:continual}, Line 1). Throughout the round, R2-VLA follows the \textsc{Act}/\textsc{Ask} decision rule (Algorithm~\ref{alg:continual}, Lines 5--8). Whenever \textsc{Ask} is triggered, the external expert provides a corrective action \(a_t^{\mathrm E}\) for the current input \(x_t\). All corrections collected during round \(q\) are stored in a temporary correction buffer \(\mathcal B_{\mathrm{corr}}^{(q)}=\{(x_m,a_m^{\mathrm E})\}_{m=1}^{N_{\mathrm{corr}}^{(q)}}\) (Algorithm~\ref{alg:continual}, Line 8). At the end of the round, these corrections are incorporated into the retrieval memory as \(\mathcal D_{\mathrm{ret}}^{(q+1)}=\mathcal D_{\mathrm{ret}}^{(q)}\cup\mathcal B_{\mathrm{corr}}^{(q)}\) (Algorithm~\ref{alg:continual}, Line 11). Notice that updating the retrieval buffer changes the behavior of the R2-VLA policy, even though the model parameters remain frozen. For the same query \(x_t\), the updated memory \(\mathcal D_{\mathrm{ret}}^{(q+1)}\) may return a different retrieved context than \(\mathcal D_{\mathrm{ret}}^{(q)}\), thereby changing both the predicted action \(\hat a_t\) and its uncertainty \(\hat u_t\). Consequently, the previously calibrated threshold \(\hat\lambda^{(q)}\) is not guaranteed to preserve the target task-failure rate after the memory update. We therefore recalibrate the threshold after each memory update. Specifically, after constructing \(\mathcal D_{\mathrm{ret}}^{(q+1)}\), we draw a fresh calibration set \(\mathcal D_{\mathrm{cal}}^{(q+1)}\) and rerun the risk-control procedure in Algorithm~\ref{alg:risk-control} using the updated retrieval buffer (Algorithm~\ref{alg:continual}, Line 12). This produces a new certified threshold \(\hat\lambda^{(q+1)}\), which is then used throughout the next adaptation round.

\begin{algorithm}[t]
\caption{Continual Deployment Adaptation with Risk Recalibration}
\label{alg:continual}
\begin{algorithmic}[1]
\REQUIRE Initial retrieval buffer $\mathcal D_{\mathrm{ret}}^{(0)}$, initial calibration set $\mathcal D_{\mathrm{cal}}^{(0)}$

\STATE Obtain $\hat\lambda^{(0)}$ by running Algorithm~\ref{alg:risk-control}
with $\mathcal D_{\mathrm{ret}}^{(0)}$ and $\mathcal D_{\mathrm{cal}}^{(0)}$

\FOR{$q=0,1,\ldots, Q-1$} 
    \STATE Initialize correction buffer
    $\mathcal B_{\mathrm{corr}}^{(q)}\leftarrow\emptyset$

    \FOR{each timestep $t$}
        \IF{$\hat u_t\leq\hat\lambda^{(q)}$}
            \STATE Execute $\hat a_t$ 
        \ELSE
            \STATE Request expert assistance, execute $a_t^{\mathrm E}$ and store the resulting expert correction in $\mathcal B_{\mathrm{corr}}^{(q)}$
        \ENDIF
    \ENDFOR

    \STATE Update retrieval memory:
    $\mathcal D_{\mathrm{ret}}^{(q+1)}
    \leftarrow
    \mathcal D_{\mathrm{ret}}^{(q)}
    \cup
    \mathcal B_{\mathrm{corr}}^{(q)}$

    \STATE Draw a fresh calibration set
    $\mathcal D_{\mathrm{cal}}^{(q+1)}$ and obtain $\hat\lambda^{(q+1)}$ by running
    Algorithm~\ref{alg:risk-control}
    with $\mathcal D_{\mathrm{ret}}^{(q+1)}$
    and $\mathcal D_{\mathrm{cal}}^{(q+1)}$
\ENDFOR
\end{algorithmic}
\end{algorithm}

\subsection{Theoretical Results}
\label{sec:theory}

We first analyze why the uncertainty predictor is relevant to the objective~\eqref{eq:r2vla-objective}. Let \(R^{\mathrm E}\) denote the task-failure probability obtained by executing the expert policy throughout the episode and let \(T\) denote the number of decision steps in an episode. Recall the average ASK rate \(H(\lambda)\) and the population uncertainty-regression loss under the trajectory distribution induced by the R2-VLA policy $\pi_{\theta,\psi}^{\lambda}$ as \(\mathcal L_U(\lambda)\).

\begin{theorem}
\label{thm:risk-bound}
Under Assumptions~\ref{ass:lips}--\ref{ass:bound-t} detailed in Appendix~\ref{app:proof-theorem2}, for any threshold \(\lambda\), any \(c>1\), and the constant $L$ detailed in Appendix~\ref{app:proof-theorem2},
\begin{align*}
R(\lambda)&\leq
R^{\mathrm E}
+
L(c-1)\eta\,\mathbb E[T](1-H(\lambda))
\\
&+
\underbrace{
Lc\,\mathbb E\left[
\sum_{t=1}^{T}
\mathbb I\{\hat u_t\leq\lambda\}\hat u_t
\right]
}_{\substack{\text{Uncertainty during ACT}}} +
\underbrace{
\frac{
\mathbb E[T]\mathcal L_U(\lambda)
}{
\operatorname{SmoothL1}(0,\log c)
}
}_{\text{Uncertainty prediction accuracy}}.
\end{align*}
\end{theorem}

The proof is provided in Appendix~\ref{app:proof-theorem2}. Theorem~\ref{thm:risk-bound} decomposes the task-failure risk into several interpretable components. The first term represents the failure probability of the expert policy, while the second term arises from the numerical stabilization constant \(\eta\) and becomes negligible when \(\eta\) is sufficiently small. The remaining two terms characterize the contribution of the learned uncertainty signal to task-level success rate. Specifically, the third term measures the cumulative predicted uncertainty over autonomously executed actions, indicating that task-failure risk can be reduced when the robot acts autonomously primarily in low-uncertainty states. The final term captures uncertainty prediction error through the uncertainty-regression loss \(\mathcal L_U(\lambda)\), highlighting the importance of accurately predicting action discrepancy, particularly avoiding severe underestimation of large action errors. Together, these results provide theoretical motivation for learning an accurate uncertainty predictor and using it to determine when the VLA can act autonomously and when expert assistance is necessary. We next establish that the risk requirement can remain certified after retrieval-memory updates.

\begin{theorem}
\label{thm:continual-risk-main}
Consider adaptation rounds \(q=0,\ldots,Q\) in Algorithm~\ref{alg:continual}. At each round \(q\), let \(\hat{\lambda}^{(q)}\) be the threshold selected by Algorithm~\ref{alg:risk-control} using confidence level \(\delta_q=\frac{\delta}{Q+1}\). Suppose that, conditional on the deployment history before round \(q\), the calibration episodes are drawn independently from the round-\(q\) deployment distribution. Then \(\Pr\left( R_q(\hat{\lambda}^{(q)})\leq\epsilon, \ \forall q=0,\ldots,Q \right) \geq 1-\delta.\)
\end{theorem}
The proof is provided in Appendix~\ref{app:proof-theorem1}. Theorem~\ref{thm:continual-risk-main} shows that continual growth of the retrieval memory does not invalidate the task-level risk guarantee, provided that the threshold is recalibrated after each memory update using fresh calibration episodes. 
\section{Experiments}
\label{sec:experiments}

\begin{table*}[t]
\centering
\caption{
Task success and ASK rate on LIBERO. SR denotes success rate. 
}
\label{tab:main-libero}
\scriptsize
\setlength{\tabcolsep}{2.8pt}
\renewcommand{\arraystretch}{1.08}
\resizebox{\textwidth}{!}{%
\begin{tabular}{ll ccc ccc ccc ccc cc}
\toprule
\multirow{2}{*}{Backbone}
& \multirow{2}{*}{Method}
& \multicolumn{3}{c}{Spatial}
& \multicolumn{3}{c}{Object}
& \multicolumn{3}{c}{Goal}
& \multicolumn{3}{c}{Long}
& \multirow{2}{*}{Avg. SR $\uparrow$}
& \multirow{2}{*}{Avg. ASK $\downarrow$}
\\
\cmidrule(lr){3-5}
\cmidrule(lr){6-8}
\cmidrule(lr){9-11}
\cmidrule(lr){12-14}
&
& SR $\uparrow$ & Target SR & ASK $\downarrow$
& SR $\uparrow$ & Target SR & ASK $\downarrow$
& SR $\uparrow$ & Target SR & ASK $\downarrow$
& SR $\uparrow$ & Target SR & ASK $\downarrow$
& &
\\
\midrule

MiniVLA
& RICL
& 58.9\% & -- & 0.0\%
& 59.1\% & -- & 0.0\%
& 60.6\% & -- & 0.0\%
& 28.5\% & -- & 0.0\%
& 51.8\% & 0.0\%
\\
\rowcolor{lightblue!100}
& \textbf{R2-VLA}
& \textbf{67.1\%}\supgain{8.2} & 65.0\% & 17.0\%
& \textbf{77.4\%}\supgain{18.3} & 65.0\% & 16.9\%
& \textbf{71.0\%}\supgain{10.4} & 65.0\% & 14.4\%
& \textbf{35.7\%}\supgain{7.2} & 35.0\% & 34.9\%
& \textbf{62.8\%}\supgain{11.0} & 20.8\%
\\
\midrule

\(\pi_0\)-FAST
& RICL
& 74.5\% & -- & 0.0\%
& 72.0\% & -- & 0.0\%
& 70.4\% & -- & 0.0\%
& 34.2\% & -- & 0.0\%
& 62.8\% & 0.0\%
\\
\rowcolor{lightblue!100}
& \textbf{R2-VLA}
& \textbf{83.2\%}\supgain{8.7} & 80.0\% & 44.6\%
& \textbf{78.9\%}\supgain{6.9} & 80.0\% & 43.2\%
& \textbf{83.1\%}\supgain{12.7} & 80.0\% & 41.7\%
& \textbf{42.7\%}\supgain{8.5} & 40.0\% & 66.3\%
& \textbf{72.0\%}\supgain{9.2} & 49.0\%
\\
\midrule

OpenVLA
& RICL
& 67.3\% & -- & 0.0\%
& 65.8\% & -- & 0.0\%
& 64.7\% & -- & 0.0\%
& 31.6\% & -- & 0.0\%
& 57.4\% & 0.0\%
\\
\rowcolor{lightblue!100}
& \textbf{R2-VLA}
& \textbf{79.3\%}\supgain{12.0} & 75.0\% & 31.8\%
& \textbf{76.5\%}\supgain{10.7} & 75.0\% & 30.1\%
& \textbf{78.4\%}\supgain{13.7} & 75.0\% & 26.8\%
& \textbf{40.2\%}\supgain{8.6} & 40.0\% & 44.6\%
& \textbf{68.6\%}\supgain{11.2} & 33.3\%
\\

\bottomrule
\end{tabular}
}
\vspace{-2ex}
\end{table*}

\begin{table*}[t]
\centering
\caption{
Comparison of expert assistance strategies on LIBERO using the $\pi_0$-FAST backbone. SR denotes success rate, and ACT denotes the average expert-action discrepancy during autonomous execution.
}
\label{tab:intervention-baselines}
\scriptsize
\setlength{\tabcolsep}{2.8pt}
\renewcommand{\arraystretch}{1.08}
\resizebox{\textwidth}{!}{%
\begin{tabular}{l cccc cccc cccc cccc cc}
\toprule
\multirow{2}{*}{Expert Assistance Strategy}
& \multicolumn{4}{c}{Spatial}
& \multicolumn{4}{c}{Object}
& \multicolumn{4}{c}{Goal}
& \multicolumn{4}{c}{Long}
& \multirow{2}{*}{Avg. SR $\uparrow$}
& \multirow{2}{*}{Avg. ASK $\downarrow$}
\\
\cmidrule(lr){2-5}
\cmidrule(lr){6-9}
\cmidrule(lr){10-13}
\cmidrule(lr){14-17}
& SR $\uparrow$ & Target SR & ASK $\downarrow$ & ACT $\downarrow$
& SR $\uparrow$ & Target SR & ASK $\downarrow$ & ACT  $\downarrow$
& SR $\uparrow$ & Target SR & ASK $\downarrow$ & ACT  $\downarrow$
& SR $\uparrow$ & Target SR & ASK $\downarrow$ & ACT $\downarrow$
& &
\\
\midrule

RICL (no help)
& 74.5\% & -- & 0.0\% & --
& 72.0\% & -- & 0.0\% & --
& 70.4\% & -- & 0.0\% & --
& 34.2\% & -- & 0.0\% & --
& 62.8\% & 0.0\%
\\

Random ASK
& 76.7\% & 80.0\% & 44.6\% & 2.64
& 70.8\% & 80.0\% & 43.2\% & 2.79
& 77.4\% & 80.0\% & 41.7\% & 2.43
& 37.1\% & 40.0\% & 66.3\% & 2.98
& 65.5\% & 49.0\%
\\

DINO Distance + Risk Control
& 80.3\% & 80.0\% & 55.3\% & 2.31
& 72.6\% & 80.0\% & 53.7\% & 2.28
& 78.4\% & 80.0\% & 51.5\% & 2.36
& 40.9\% & 40.0\% & 77.9\% & 2.69
& 68.1\% & 59.6\%
\\

Action Entropy + Risk Control
& 80.2\% & 80.0\% & 54.9\% & 2.01
& 77.8\% & 80.0\% & 53.1\% & 2.15
& 81.5\% & 80.0\% & 51.0\% & 1.96
& 41.4\% & 40.0\% & 77.2\% & 2.23
& 70.2\% & 59.1\%
\\

\rowcolor{lightblue!100}
\textbf{Learned Uncertainty + Risk Control (ours)}
& \textbf{83.2\%} & 80.0\% & 44.6\% & 1.58
& \textbf{78.9\%} & 80.0\% & 43.2\% & 1.66
& \textbf{83.1\%} & 80.0\% & 41.7\% & 1.46
& \textbf{42.7\%} & 40.0\% & 66.3\% & 1.70
& \textbf{72.0\%} & \textbf{49.0\%}
\\

\bottomrule
\end{tabular}
}
\vspace{-2ex}
\end{table*}

\begin{figure*}[t]
    \centering
    \includegraphics[width=\linewidth] 
    {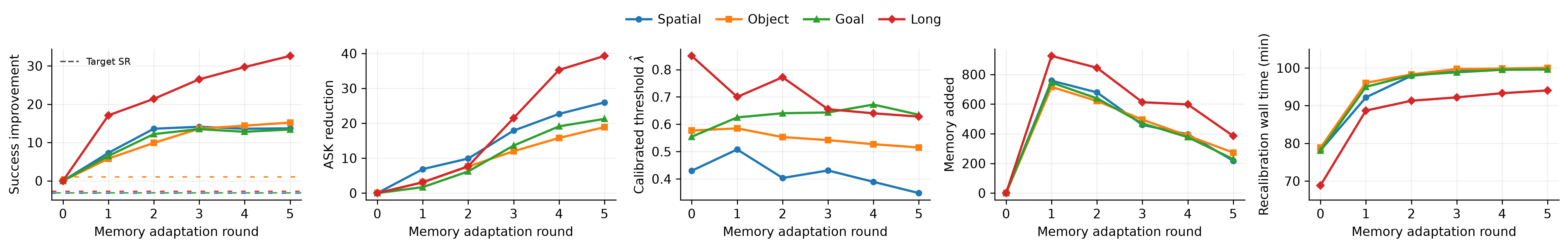}
    \caption{
    Continual memory adaptation of R2-VLA using the $\pi_0$-FAST backbone across four LIBERO suites. From left to right: improvement in task success rate relative to the initial policy, reduction in ASK rate, calibrated threshold $\hat{\lambda}$, number of expert corrections added to retrieval memory, and recalibration wall time.
}
    \label{fig:continual}
    \vspace{-2ex}
\end{figure*}

%We design our experiments to evaluate the proposed R2-VLA framework regarding the following research questions: 1. Does the proposed risk-control procedure achieve the user-specified target task success rate? 2. Does the learned uncertainty identify states that require external assistance more effectively then similarity based method? 3. As the retrieval memory growth, can the robot maintain the target reliability? 

\subsection{Experimental Setup}
\label{sec:exp-setup}

We implement R2-VLA policy following the model architecture adopted in RICL~\cite{sridhar2025ricl}, which uses $\pi_0$-FAST as the VLA backbone. To evaluate the generality of our method across different model architectures and scales, we additionally consider MiniVLA~\cite{belkhale2024minivla} and OpenVLA~\cite{kim2024openvla} as smaller- and larger-scale VLA models, respectively. We evaluate R2-VLA on LIBERO~\cite{liu2023libero}, a simulated robotic manipulation benchmark based on the Franka Panda arm. LIBERO contains four suites designed to evaluate different forms of manipulation generalization: \textsc{Spatial}, \textsc{Object}, \textsc{Goal}, and \textsc{Long}. For each suite, we use 20 expert demonstrations per task to construct the retrieval buffer and maintain a disjoint set of 20 demonstrations per task for risk calibration. During evaluation, when \textsc{Ask} is triggered, a separately trained expert policy provides the action for the current observation. We report task success over 100 evaluation rollouts per task using a held-out evaluation set that is disjoint from both the retrieval and calibration buffers. In addition to task success rate, we report the ASK rate, denoted by \(\widehat H(\lambda)\) as defined in Section~\ref{sec:risk-control}, which measures the fraction of decision points at which R2-VLA requests expert assistance. Additional implementation details are provided in Appendix~\ref{app:implement}. Additional ablation studies investigating the effects of the target risk level \(\epsilon\), retrieval corruption during uncertainty-head training, and calibration-set size \(N_{\mathrm{cal}}\) are provided in Appendix~\ref{app:ablation}.

\subsection{Results}
\label{sec:exp-main}

Table~\ref{tab:main-libero} reports the task success rates and ASK rate of R2-VLA compared with RICL. R2-VLA consistently improves task success over the corresponding RICL baseline across all backbone--task suite combinations, demonstrating the effectiveness of identifying uncertain retrieval-guided behaviors and requesting expert assistance. Moreover, R2-VLA generally achieves the prescribed task-level success targets, which are typically set at least \(5\) percentage points above the corresponding RICL baseline. Specifically, R2-VLA reaches or exceeds the target success rate in 11 of the 12 settings. The only exception is \(\pi_0\)-FAST on LIBERO-\textsc{Object}, where the achieved success rate of \(78.9\%\) is slightly below the \(80.0\%\) target. This small diviation may be due to limited number of evaluation rollouts, as our risk-control guarantee in Theorem~\ref{thm:continual-risk-main} is statistical rather than a deterministic guarantee on every evaluation set. The ASK rate further reflect the varying difficulty of successful execution across task suites. In particular, LIBERO-\textsc{Long}, which exhibits the lowest success rate across all three backbones, also requires the highest ASK rate. In contrast, the other suites generally require fewer expert assistance while still achieving their prescribed success targets. These results suggest that R2-VLA can adapt its reliance on expert assistance to the uncertain of predicted execution across different tasks and VLA architectures.

Improved task success alone does not establish that the learned uncertainty identifies decision points where retrieved demonstrations are insufficient for successful VLA execution, as task success may also improve simply by delegating more actions to an expert. We therefore compare R2-VLA against a \textbf{Random ASK} baseline that requests expert assistance at randomly selected decision points while matching the expert assistance frequency of our method. This comparison isolates the benefit of identifying uncertain retrieval-guided behaviors rather than merely increasing expert involvement. To further investigate whether learning action uncertainty from the VLA is necessary, we consider two alternative uncertainty signals: (1) \textbf{DINO Distance}, which uses the visual feature distance employed by RICL for demonstration retrieval as its uncertainty score, and (2) \textbf{Action Entropy}, which instead estimates uncertainty from the VLA's action distribution. The DINO Distance variant directly evaluates whether the original retrieval quality metric is sufficient to identify when retrieved demonstrations provide insufficient behavioral guidance, while the Action Entropy variant evaluates whether the model's predictive uncertainty can identify uncertain actions without training an additional uncertainty estimator. For these two baselines, thresholds are independently calibrated using the same risk-control procedure as R2-VLA, allowing us to compare the resulting trade-offs between task success and ASK rate. We additionally report the average action discrepancy during ACT, defined as \(\frac{\mathbb E\left[\sum_{t=1}^{T}\mathbb I\{\mathrm{ACT}_t\}\hat u_t\right]}{\mathbb E\left[\sum_{t=1}^{T}\mathbb I\{\mathrm{ACT}_t\}\right]}\). This metric provides empirical support for the mechanism highlighted in Theorem~\ref{thm:risk-bound} by evaluating whether R2-VLA effectively prevents actions with large expert discrepancies from being executed autonomously.

Table~\ref{tab:intervention-baselines} shows that R2-VLA outperforms all three baselines. Specifically, under the same ASK budget, Random ASK achieves an average success rate that is \(6.5\) percentage points lower than R2-VLA. This demonstrates that the improvement cannot be attributed solely to additional expert assistance. Instead, selectively identifying decision points where retrieval-guided behaviors are uncertain is critical for successful task execution. Compared with DINO Distance, R2-VLA achieves higher task success while requiring fewer expert assistance. This indicates that retrieval quality metrics alone are insufficient to determine whether the retrieved demonstrations enable the VLA to successfully complete the task, as illustrated in Fig.~\ref{fig:qualitative}(a). Similarly, R2-VLA outperforms Action Entropy, suggesting that uncertainty derived directly from the action distribution is less effective than our learned uncertainty predictor. Notably, R2-VLA achieves a more favorable empirical trade-off between task success and ASK rate than both DINO Distance and Action Entropy. Moreover, R2-VLA achieves the lowest ACT error across all four LIBERO suites. This further supports Theorem~\ref{thm:risk-bound}, showing that the learned uncertainty effectively identifies actions for autonomous execution, thereby contributing to lower task-level failure risk. Overall, these results demonstrate the effectiveness of learning action uncertainty from retrieval-augmented VLA representations. Complete results for other VLA backbones are provided in Appendix~\ref{app:intervention-baseline} Table~\ref{tab:intervention-baselines-additional}.

We next evaluate whether expert corrections collected through \textsc{Ask} can improve future task success while reducing the need for external assistance. Following Algorithm~\ref{alg:continual}, we divide deployment into adaptation rounds. During each round, the retrieval buffer and calibrated threshold \(\hat\lambda\) remain fixed, and expert corrections are collected whenever the robot requests assistance. At the end of each round, these corrections are incorporated into the retrieval memory, followed by recalibration using Algorithm~\ref{alg:risk-control}. Specifically, we execute fresh calibration using the updated retrieval memory as detailed in Appendix~\ref{app:benchmark-eval}, ensuring that calibration satisfies the requirement of Theorem~\ref{thm:continual-risk-main}. For consistent comparison across rounds, we evaluate each updated policy on the same held-out evaluation sets as before. We maintain the same fixed target success rate as in Table~\ref{tab:main-libero}.

Figure~\ref{fig:continual} illustrates the evolution of R2-VLA using the $\pi_0$-FAST backbone across memory adaptation rounds. As expert corrections are progressively incorporated into the retrieval memory, task success consistently improves while the frequency of expert assistance decreases. Importantly, after the initial memory update, the observed success rates meet or exceed the prescribed targets across adaptation rounds, providing empirical evidence consistent with Theorem~\ref{thm:continual-risk-main}. Meanwhile, the decreasing ASK rate suggests that accumulated expert corrections provide increasingly useful retrieval experiences, enabling the VLA to achieve higher task success with less external assistance. We further examine how the calibrated threshold $\hat{\lambda}$ evolves as the retrieval memory grows. As shown in Figure~\ref{fig:continual}, the threshold varies across adaptation rounds, with distinct patterns across task suites. These changes reflect the evolving distribution of uncertainty scores induced by updates to the retrieval memory, which can alter both the retrieved demonstrations and the resulting action predictions. Therefore, a threshold calibrated for an earlier retrieval buffer may no longer provide the desired risk control after memory updates. By recalibrating $\hat{\lambda}$ after each adaptation round, R2-VLA adjusts its ACT/ASK decisions to the updated policy. The results also show that this recalibration can be performed with relatively modest computational overhead, with calibration wall time remaining approximately stable across later adaptation rounds. Together, these findings demonstrate that R2-VLA can progressively improve autonomous task execution and reduce expert dependence while empirically maintaining the prescribed task-level success rate requirement.

\section{Conclusion}
In this work, we proposed R2-VLA, a risk-controlled retrieval-augmented VLA framework that enables reliable test-time adaptation by satisfying a user-specified task success rate requirement with minimal expert assistance. R2-VLA learns an uncertainty predictor to determine when to act autonomously and when to request expert assistance. We further introduced a risk-control procedure that calibrates the uncertainty threshold to satisfy the prescribed task-level failure risk requirement. Moreover, R2-VLA supports continual adaptation by incorporating expert corrections into retrieval memory and preserving task-level risk control. Experiments across various settings demonstrate consistent improvements in task success over RICL while requiring less frequent expert assistance. Overall, our work represents a step toward building more reliable and adaptive robotic systems that can recognize their own limitations, seek assistance when necessary, and continually improve through interaction with their environments.

\section*{LLM Usage Statement}
OpenAI's ChatGPT was used to assist with language editing, improving the organization of the manuscript, and refining the presentation. All research contributions, experimental results, and scientific claims were reviewed and validated by the authors, who take full responsibility for the content of this paper.
{
    \small
    \bibliographystyle{ieeenat_fullname}
    \bibliography{main}
}

\clearpage
\setcounter{page}{1}
\maketitlesupplementary

\section{Related Work}
\label{app:related-work}

\noindent\textbf{Retrieval-Augmented VLA.}
While recent vision-language-action (VLA) models, such as OpenVLA~\cite{kim2024openvla} and $\pi_0$~\cite{black2024pi_0}, have demonstrated strong generalization through pretraining on large and diverse robot datasets, adapting them to new tasks often still requires task-specific fine-tuning or retraining, limiting their ability to behave as truly generalist agents. In contrast, humans can rapidly adapt to novel tasks by drawing on prior experience~\cite{baddeley2006working,robertson2012new,tulving1972episodic}. Motivated by this capability, recent work has explored in-context imitation learning, in which demonstrations of a new task are provided directly to the VLA at inference time, enabling adaptation without updating the model parameters. ICRT~\cite{fu2024context} is an early representative approach, formulating robot control as next-token prediction over sequences of visual observations, proprioceptive states, and actions. By placing one or more human teleoperation trajectories in the transformer's context, ICRT can infer and execute previously unseen tasks without test-time parameter updates, demonstrating that the in-context learning behavior of autoregressive models can extend from language to continuous robotic control. More recently, RICL~\cite{sridhar2025ricl} showed that such capability does not naturally emerge in pretrained VLAs and proposed explicitly post-training a pretrained \(\pi_0\)-FAST policy to exploit retrieved demonstrations. At deployment, RICL retrieves relevant states and actions from a small demonstration buffer and incorporates them into the model context, enabling adaptation to new manipulation tasks using roughly 10--20 demonstrations without updating model parameters.

A parallel direction investigates whether existing frozen VLAs can benefit from demonstrations without additional in-context learning (ICL) training. Retrieval-VLA~\cite{zhang2026retrieval} enables training-free adaptation by identifying relevant past experiences and directly incorporating the retrieved information into the predictions of a frozen VLA. Since the VLA does not need to learn how to interpret retrieved demonstrations as in-context examples, this approach avoids ICL-specific post-training. Similarly, ReCAP~\cite{park2026retrieve} extends VLA policies to new tasks by conditioning on demonstrations from an external retrieval pool without task-specific retraining. Beyond directly leveraging retrieved demonstrations, recent methods further explore how to improve their relevance and utilization. StellaVLA~\cite{xu2026stellavla} converts retrieved trajectories into structured task plans, subgoals, and motion descriptions to provide richer task-specific guidance. RA-VLA~\cite{jang2026ra} introduces behavior-aligned retrieval and grounded execution to improve the relevance of retrieved demonstrations, while ICI-VLA~\cite{yang2026ici} employs spatiotemporally aligned demonstrations to better match the current task phase and geometry. Collectively, these studies demonstrate the potential of retrieval-based adaptation for extending VLA capabilities without task-specific parameter updates. Building on these advances, our work addresses the complementary challenge of ensuring successful task execution given retrieved demonstrations by introducing uncertainty-guided expert assistance and task-level risk control.

\noindent\textbf{Uncertainty in Robotics.}
Uncertainty has long been used in robot learning to determine when autonomous execution of the robot should be deferred to an expert~\cite{chernova2007confidence,hoque2021thriftydagger,menda2019ensembledagger,zhao2025conformalized}. Early confidence-based learning-from-demonstration methods allow a robot to request demonstrations in states where its learned policy is insufficiently confident~\cite{chernova2007confidence}. This principle was later incorporated into interactive imitation learning. EnsembleDAgger~\cite{menda2019ensembledagger} uses ensemble disagreement to determine when a learned policy can safely act~\cite{menda2019ensembledagger}.  More recently, ConformalDAgger~\cite{zhao2025conformalized} replaces heuristic uncertainty thresholds with online conformal calibration, enabling uncertainty estimates to adapt under deployment-time expert shift. 

Recent work has extended uncertainty estimation to VLA models. SAFE~\cite{gu2026safe} estimates uncertainty from the internal representations of a VLA to assess the reliability of ongoing task execution and determine when intervention may be needed. ReconVLA~\cite{chen2026reconvla} uses conformalized action uncertainty to compare candidate actions and select the least uncertain action for execution.~\cite{romer2026uncertainty} estimate epistemic uncertainty in flow-based VLAs to identify unreliable predictions and prioritize informative samples for active adaptation. SCALE~\cite{choi2026scale} uses self-uncertainty to adapt both visual perception and action generation, encouraging broader exploration under high uncertainty and more confident execution under low uncertainty.~\cite{ghaleb2026uncertainty} use calibrated uncertainty to gate control among normal VLA execution, pause-and-reobservation, and fallback planning. Collectively, these methods demonstrate the value of uncertainty estimation for identifying potentially incorrect robot actions and determining when expert assistance is necessary. Building on these advances, our work introduces task-level risk control to calibrate the uncertainty threshold, enabling the VLA to satisfy a user-specified task success requirement while minimizing the frequency of expert assistance.

\noindent\textbf{Risk Control for Robotics.}
Statistical risk-control techniques~\cite{bates2021distribution,angelopoulos2025learn,angelopoulos2024conformal} have recently been applied to LLM-based robotic planning. KnowNo~\cite{ren2023robots} uses conformal prediction to construct sets of candidate robot actions, executing autonomously when the prediction set is sufficiently unambiguous and otherwise requesting human assistance. Introspective Planning~\cite{liang2024introspective} improves the underlying LLM reasoning before conformal calibration to reduce unnecessary clarification while retaining statistical success guarantees.~\cite{wang2024probabilistically} extends conformal planning to multi-robot systems.~\cite{wang2025cofinellm} propose conformal-aware finetuning to reduce prediction-set size and human-help rates. In contrast, we study risk control for VLAs during closed-loop task execution. Rather than calibrating the uncertainty of an individual action, we calibrate the uncertainty threshold that determines when the VLA acts autonomously and when it requests expert assistance. The threshold is evaluated through complete task rollouts, so the resulting risk criterion directly reflects whether the induced VLA policy succeeds or fails at the task level. This formulation also accounts for the sequential nature of VLA execution, where locally uncertain decisions can affect subsequent states and ultimately determine task success.

\section{Proofs}
\label{app:proofs}

\subsection{Proof of Theorem~\ref{thm:risk-bound}}
\label{app:proof-theorem2}

Before proving Theorem~\ref{thm:risk-bound}, we first recall several definitions from the main paper. Let \(a_t^{\mathrm E}\) denote the expert action chunk at decision step \(t\), and let \(R^{\mathrm E}\) denote the task-failure probability when the expert policy is executed throughout the episode. For the VLA prediction \(\hat a_t\), recall that
\[
u_t
=
d_A(\hat a_t,a_t^{\mathrm E})
\]
denotes the discrepancy between the VLA-predicted action and the corresponding expert action. Let \(T\) denote the number of decision steps in an episode. For a calibrated threshold \(\lambda\), we define the population ASK rate as
\[
H(\lambda)
=
\frac{
\mathbb E
\left[
\sum_{t=1}^{T}
\mathbb I\{\hat u_t>\lambda\}
\right]
}{
\mathbb E[T]
}.
\]
We further define
\[
\operatorname{SmoothL1}(x,y)
=
\begin{cases}
\frac{1}{2}(x-y)^2,
&
|x-y|\leq 1,
\\[1mm]
|x-y|-\frac{1}{2},
&
|x-y|>1,
\end{cases}
\]
and let
\[
\mathcal L_U(\lambda)
=
\frac{
\mathbb E
\left[
\sum_{t=1}^{T}
\operatorname{SmoothL1}
\left(
\log(\hat u_t+\eta),
\log(u_t+\eta)
\right)
\right]
}{
\mathbb E[T]
}
\]
denote the population uncertainty-regression loss under the trajectory distribution induced by the R2-VLA policy with threshold \(\lambda\).

We next introduce the assumptions used to establish the risk bound.
\begin{assumption}
\label{ass:lips}
Let
\[
P(\cdot\mid x,a)
\]
denote the distribution of the next decision state after executing an
action chunk \(a\) from state \(x\). There exists \(L>0\) such that,
for any state \(x\) and action chunks \(a,a'\),
\[
\left\|
P(\cdot\mid x,a)
-
P(\cdot\mid x,a')
\right\|_{\mathrm{TV}}
\leq
L \,d_A(a,a').
\]
\end{assumption}

Assumption~\ref{ass:lips} is a standard smoothness condition for continuous-control systems. It requires the environment dynamics to vary continuously with the executed action. If two action chunks are close, then the corresponding next-state distributions are also close. Similar Lipschitz-type assumptions on transition dynamics are commonly used in reinforcement learning~\cite{bennett2024lowrankmdpscontinuousaction,chen2024robust,liu2026robustgeneralutilityreinforcement}.

\begin{assumption}
\label{ass:bound-t}
The expected number of decision steps is finite:
\[
\mathbb E[T]<\infty.
\]
\end{assumption}

Assumption~\ref{ass:bound-t} is naturally satisfied in finite-horizon robotic manipulation tasks, where each episode terminates after task completion, failure, or a predefined maximum number of decision steps. We next prove Theorem~\ref{thm:risk-bound}.

\begin{proof}
Fix \(c>1\). Consider a decision step \(t\) at which the robot acts autonomously, i.e.,
\[
\hat u_t\leq\lambda.
\]
We separate two cases according to whether the learned uncertainty substantially underestimates the true action error. First, suppose
\[
u_t+\eta
\leq
c(\hat u_t+\eta).
\]
Rearranging gives
\[
u_t
\leq
c \hat u_t+(c-1)\eta.
\]
By definition,
\[
u_t
=
d_A(\hat a_t,a_t^{\mathrm E}),
\]
and therefore
\[
d_A(\hat a_t,a_t^{\mathrm E})
\leq
c \hat u_t+(c-1)\eta.
\]
Applying Assumption~\ref{ass:lips} with \(a=\hat a_t\) and \(a'=a_t^{\mathrm E}\), we obtain
\[
\begin{aligned}
\left\|
P(\cdot\mid x_t,\hat a_t)
-
P(\cdot\mid x_t,a_t^{\mathrm E})
\right\|_{\mathrm{TV}}
&\leq
L\,d_A(\hat a_t,a_t^{\mathrm E})
\\
&\leq
L\left[
c \hat u_t+(c-1)\eta
\right].
\end{aligned}
\]
Now consider the complementary case
\[
u_t+\eta
>
c(\hat u_t+\eta).
\]
Since \(c>1\), taking logarithms gives
\[
\log(u_t+\eta)
-
\log(\hat u_t+\eta)
>
\log c
>
0.
\]
Hence,
\[
\left|
\log(u_t+\eta)
-
\log(\hat u_t+\eta)
\right|
>
\log c.
\]
Recall that the SmoothL1 loss depends only on the magnitude of the residual:
\[
\operatorname{SmoothL1}(x,y)
=
\begin{cases}
\frac{1}{2}|x-y|^2, & |x-y|\leq 1,\\[1mm]
|x-y|-\frac{1}{2}, & |x-y|>1.
\end{cases}
\]
Therefore, a larger absolute residual always produces a no smaller SmoothL1 loss. Since
\[
\left|
\log(u_t+\eta)
-
\log(\hat u_t+\eta)
\right|
>
\log c,
\]
we have
\[
\operatorname{SmoothL1}
\left(
\log(\hat u_t+\eta),
\log(u_t+\eta)
\right)
\geq
\operatorname{SmoothL1}(0,\log c).
\]
Hence,
\begin{align*}
&\mathbb I
\left\{
u_t+\eta>c(\hat u_t+\eta)
\right\} \\
&\quad\leq
\frac{
\operatorname{SmoothL1}
\left(
\log(\hat u_t+\eta),
\log(u_t+\eta)
\right)
}{
\operatorname{SmoothL1}(0,\log c)
}.
\end{align*}
Then for an autonomous decision \(\hat u_t\leq\lambda\), decompose the transition discrepancy according to whether the uncertainty substantially underestimates the action error:
\begin{align*}
&
\left\|
P(\cdot\mid x_t,\hat a_t)
-
P(\cdot\mid x_t,a_t^{\mathrm E})
\right\|_{\mathrm{TV}}
\\
&=
\left\|
P(\cdot\mid x_t,\hat a_t)
-
P(\cdot\mid x_t,a_t^{\mathrm E})
\right\|_{\mathrm{TV}}
\mathbb I
\left\{
u_t+\eta
\leq
c(\hat u_t+\eta)
\right\}
\\
&+
\left\|
P(\cdot\mid x_t,\hat a_t)
-
P(\cdot\mid x_t,a_t^{\mathrm E})
\right\|_{\mathrm{TV}}
\mathbb I
\left\{
u_t+\eta
>
c(\hat u_t+\eta)
\right\}.
\end{align*}
For the first term, we have already shown that
\[
u_t+\eta
\leq
c(\hat u_t+\eta)
\]
implies
\[
\left\|
P(\cdot\mid x_t,\hat a_t)
-
P(\cdot\mid x_t,a_t^{\mathrm E})
\right\|_{\mathrm{TV}}
\leq
L\left[
c \hat u_t+(c-1)\eta
\right].
\]
Therefore,
\begin{align*}
&
\left\|
P(\cdot\mid x_t,\hat a_t)
-
P(\cdot\mid x_t,a_t^{\mathrm E})
\right\|_{\mathrm{TV}}
\mathbb I
\left\{
u_t+\eta
\leq
c(\hat u_t+\eta)
\right\}
\\
&\qquad\leq
L\left[
c \hat u_t+(c-1)\eta
\right].
\end{align*}
For the second term, total variation distance is always bounded by one, so
\begin{align*}
&
\left\|
P(\cdot\mid x_t,\hat a_t)
-
P(\cdot\mid x_t,a_t^{\mathrm E})
\right\|_{\mathrm{TV}}
\mathbb I
\left\{
u_t+\eta
>
c(\hat u_t+\eta)
\right\}
\\
&\qquad\leq
\mathbb I
\left\{
u_t+\eta
>
c(\hat u_t+\eta)
\right\}
\\
&\qquad\leq
\frac{
\operatorname{SmoothL1}
\left(
\log(\hat u_t+\eta),
\log(u_t+\eta)
\right)
}{
\operatorname{SmoothL1}(0,\log c)
}.
\end{align*}
Combining the two cases therefore gives, whenever
\(\hat u_t\leq\lambda\),
\begin{align*}
&
\left\|
P(\cdot\mid x_t,\hat a_t)
-
P(\cdot\mid x_t,a_t^{\mathrm E})
\right\|_{\mathrm{TV}}
\\
&\leq
L\left[
c \hat u_t+(c-1)\eta
\right]
+
\frac{
\operatorname{SmoothL1}
\left(
\log(\hat u_t+\eta),
\log(u_t+\eta)
\right)
}{
\operatorname{SmoothL1}(0,\log c)
}.
\end{align*}
We now lift the one-step transition bound to the entire trajectory. Consider a coupling of two trajectories that start from the same initial state: one generated by the R2-VLA policy and one generated by the expert policy. Let
\[
A
=
\left\{
\text{the two trajectories differ at some decision step}
\right\}
\]
denote the event that the two trajectories ever separate. If \(A\) does not occur, the two trajectories are identical throughout the episode and therefore have the same task outcome. Hence, if the R2-VLA policy trajectory fails, then either the expert trajectory also fails or the two trajectories must have separated. Therefore,
\[
\left\{
\text{R2-VLA policy failure}
\right\}
\subseteq
\left\{
\text{expert-policy failure}
\right\}
\cup A,
\]
which implies
\[
R(\lambda)
\leq
R^{\mathrm E}
+
\Pr(A).
\]
It therefore remains to bound the probability that the two trajectories separate. Let \(A_t\) denote the event that the trajectories agree up to decision step \(t\), but their next decision states differ after the action at step \(t\). The events \(A_1,\ldots,A_T\) are disjoint because each trajectory pair has a unique first separation time. Thus,
\[
\Pr(A)
=
\mathbb E
\left[
\sum_{t=1}^{T}
\mathbb I\{A_t\}
\right].
\]
Conditioned on the two trajectories having remained identical up to decision step \(t\), they are at the same state \(x_t\). If \(\hat u_t>\lambda\), the R2-VLA policy executes the expert action \(a_t^{\mathrm E}\), so the two policies induce identical next-state distributions and cannot separate at that step. If \(\hat u_t\leq\lambda\), the R2-VLA policy executes \(\hat a_t\). Under a maximal coupling of the two next-state distributions, the conditional probability that the next states differ is bounded by
\[
\left\|
P(\cdot\mid x_t,\hat a_t)
-
P(\cdot\mid x_t,a_t^{\mathrm E})
\right\|_{\mathrm{TV}}.
\]
Consequently,
\[
\Pr(A)
\leq
\mathbb E
\left[
\sum_{t=1}^{T}
\mathbb I\{\hat u_t\leq\lambda\}
\left\|
P(\cdot\mid x_t,\hat a_t)
-
P(\cdot\mid x_t,a_t^{\mathrm E})
\right\|_{\mathrm{TV}}
\right].
\]
Using the one-step bound derived above for every autonomous decision,
\begin{align*}
&\Pr(A)
\leq
Lc\,
\mathbb E
\left[
\sum_{t=1}^{T}
\mathbb I\{\hat u_t\leq\lambda\}\hat u_t
\right]
\\
&+
L(c-1)\eta\,
\mathbb E
\left[
\sum_{t=1}^{T}
\mathbb I\{\hat u_t\leq\lambda\}
\right]
\\
&+
\frac{
\mathbb E
\left[
\sum_{t=1}^{T}
\mathbb I\{\hat u_t\leq\lambda\}
\operatorname{SmoothL1}
\left(
\log(\hat u_t+\eta),
\log(u_t+\eta)
\right)
\right]
}{
\operatorname{SmoothL1}(0,\log c)
}.
\end{align*}
Combining this inequality with
\[
R(\lambda)
\leq
R^{\mathrm E}
+
\Pr(A)
\]
gives
\begin{align*}
&R(\lambda)
\leq\;
R^{\mathrm E}
+
Lc\,
\mathbb E
\left[
\sum_{t=1}^{T}
\mathbb I\{\hat u_t\leq\lambda\}\hat u_t
\right]
\\
&+
L(c-1)\eta\,
\mathbb E
\left[
\sum_{t=1}^{T}
\mathbb I\{\hat u_t\leq\lambda\}
\right]
\\
&+
\frac{
\mathbb E
\left[
\sum_{t=1}^{T}
\mathbb I\{\hat u_t\leq\lambda\}
\operatorname{SmoothL1}
\left(
\log(\hat u_t+\eta),
\log(u_t+\eta)
\right)
\right]
}{
\operatorname{SmoothL1}(0,\log c)
}.
\end{align*}
By the definition of \(H(\lambda)\),
\[
\mathbb E
\left[
\sum_{t=1}^{T}
\mathbb I\{\hat u_t\leq\lambda\}
\right]
=
\mathbb E[T]
\left(
1-H(\lambda)
\right).
\]
Moreover,
\[
\begin{aligned}
&
\mathbb E
\left[
\sum_{t=1}^{T}
\mathbb I\{\hat u_t\leq\lambda\}
\operatorname{SmoothL1}
\left(
\log(\hat u_t+\eta),
\log(u_t+\eta)
\right)
\right]
\\
&\qquad\leq
\mathbb E
\left[
\sum_{t=1}^{T}
\operatorname{SmoothL1}
\left(
\log(\hat u_t+\eta),
\log(u_t+\eta)
\right)
\right]
\\
&\qquad=
\mathbb E[T]\mathcal L_U(\lambda).
\end{aligned}
\]
Substituting these two relations gives
\begin{align*}
&R(\lambda)
\leq\;
R^{\mathrm E}
+
Lc\,
\mathbb E
\left[
\sum_{t=1}^{T}
\mathbb I\{\hat u_t\leq\lambda\}\hat u_t
\right]
\\
&\quad+
L(c-1)\eta\,
\mathbb E[T]
\left(
1-H(\lambda)
\right)
+
\frac{
\mathbb E[T]\mathcal L_U(\lambda)
}{
\operatorname{SmoothL1}(0,\log c)
}.
\end{align*}
This completes the proof.
\end{proof}

Theorem~\ref{thm:risk-bound} connects the uncertainty-training objective directly to the task-level failure risk of the R2-VLA policy. The bound contains three sources of task failure beyond the expert risk \(R^{\mathrm E}\). The first term,
\[
Lc\,
\mathbb E
\left[
\sum_{t=1}^{T}
\mathbb I\{\hat u_t\leq\lambda\}\hat u_t
\right],
\]
measures the total uncertainty accumulated over decisions that are executed autonomously and it goes to zero when the uncertainty goes to zero. The second term,
\[
L(c-1)\eta\,
\mathbb E[T]
\left(
1-H(\lambda)
\right),
\]
arises from the numerical stabilization constant \(\eta\) used in the regression objective of learning the uncertainty head. Since \(\eta\) is chosen to be very small, this term is negligible in practice and vanishes as \(c\rightarrow1\). The final term,
\[
\frac{
\mathbb E[T]\mathcal L_U(\lambda)
}{
\operatorname{SmoothL1}(0,\log c)
},
\]
captures uncertainty underestimation. A smaller population uncertainty-regression loss reduces the probability that the R2-VLA reports small uncertainty while making a large action error.

Thus, the theorem shows that the task risk approaches the expert risk when the uncertainty head accurately predicts action error and the VLA acts autonomously primarily in low-uncertainty states. In particular, suppose \(R^{\mathrm E}\rightarrow0\), \(\mathcal L_U(\lambda)\rightarrow0\), and
\[
\mathbb E
\left[
\sum_{t=1}^{T}
\mathbb I\{\hat u_t\leq\lambda\}\hat u_t
\right]
\rightarrow0,
\]
i.e., the robot becomes increasingly confident on the states where it acts autonomously, while \(\mathbb E[T]\) remains bounded. Then, by choosing \(c\rightarrow1\) sufficiently slowly, the remaining terms in the bound also vanish, implying
\[
R(\lambda)\rightarrow0.
\]

\subsection{Proof of Theorem~\ref{thm:continual-risk-main} }
\label{app:proof-theorem1}

Before proving Theorem~\ref{thm:continual-risk-main}, we first establish that, within each adaptation round, the threshold selected by Algorithm~\ref{alg:risk-control} satisfies the desired task-level success requirement.
\begin{theorem}
\label{thm:single-risk}
Consider a fixed retrieval buffer \(\mathcal D_{\mathrm{ret}}\) and a finite candidate set \(\Lambda=\{\lambda_1,\ldots,\lambda_K\}\). Suppose the \(N_{\mathrm{cal}}\) calibration episodes are drawn independently from the deployment distribution. Let \(\widehat{\Lambda}_{\mathrm{safe}}\) be the set of thresholds certified by Algorithm~\ref{alg:risk-control} and \(\hat\lambda\) be the threshold returned by  Algorithm~\ref{alg:risk-control}. Then, with probability at least \(1-\delta\),
\[
R(\hat\lambda)\leq\epsilon.
\]
More strongly,
\[
\Pr\left(
R(\lambda)\leq\epsilon,
\ \forall\lambda\in\widehat{\Lambda}_{\mathrm{safe}}
\right)
\geq
1-\delta.
\]
\end{theorem}

\begin{proof}
Consider any candidate threshold \(\lambda\) whose population task-failure risk exceeds the prescribed target, i.e.,
\[
R(\lambda)>\epsilon.
\]
The episode-level losses
\[
L_1(\lambda),\ldots,L_{N_{\mathrm{cal}}}(\lambda)
\]
are independent Bernoulli random variables with failure probability \(R(\lambda)\). Hence,
\[
F_{\lambda}
\sim
\operatorname{Binomial}
\left(
N_{\mathrm{cal}},
R(\lambda)
\right),
\]
where \(F_\lambda=\sum_{i=1}^{N_{\mathrm{cal}}}L_i(\lambda)\) is the number of failed episodes. We then ask whether the observed failure count \(F_\lambda\) is sufficiently small to reject the possibility that the task-failure risk \(R(\lambda)\) exceeds the prescribed target \(\epsilon\). Following~\cite{angelopoulos2025learn}, we formulate this as the hypothesis test
\[
 \mathcal H_{0,\lambda}: R(\lambda)>\epsilon.
\]
To certify \(\lambda\), we therefore need to show that the observed number of calibration failures is unusually small under this null hypothesis. Suppose first that the true failure rate were exactly equal to the target \(\epsilon\). Then the number of failures among \(N_{\mathrm{cal}}\) independent calibration episodes would follow
\[
B\sim\operatorname{Binomial}(N_{\mathrm{cal}},\epsilon).
\]
For the observed failure count \(F_{\lambda}\), we compute
\[
p_{\lambda}
=
\Pr\left(
B\leq F_{\lambda}
\right).
\]
This is the probability of observing \(F_{\lambda}\) or fewer failures when the failure rate is \(\epsilon\). If the threshold actually makes \(R(\lambda)>\epsilon,\) then failures would occur more frequently than under the boundary case \(R(\lambda)=\epsilon\). For example, suppose our target failure rate is \(10\%\), so \(\epsilon=0.1\), and we observe zero failures in \(50\) calibration episodes. If the true failure rate were exactly \(10\%\),
\[
\Pr(F=0)=0.9^{50}\approx 0.0052.
\]
If the threshold were actually worse, say its true failure rate were \(20\%\),
\[
\Pr(F=0)=0.8^{50}\approx 1.4\times 10^{-5}.
\]
Thus, among all failure rates \(R(\lambda)>\epsilon\), the case closest to \(\epsilon\) makes an unusually favorable calibration outcome, such as ``zero failures,'' most likely. Consequently, observing the same small number \(F_{\lambda}\) of failures would be even less likely under a larger failure probability. Therefore,
\[
\Pr_{B\sim\operatorname{Binomial}(N_{\mathrm{cal}},R(\lambda))}
\left(
B\leq F_{\lambda}
\right)
\leq
p_{\lambda}.
\]
Thus, evaluating the probability at the boundary value \(\epsilon\) deliberately gives the null hypothesis its most favorable case. The resulting \(p_{\lambda}\) may therefore be larger than the exact probability under the true failure rate, making it harder to certify \(\lambda\). In this sense, \(p_{\lambda}\) is a conservative \(p\)-value for testing whether
\[
R(\lambda)>\epsilon.
\]
Our procedure in Algorithm~\ref{alg:risk-control} certifies \(\lambda\) only when
\[
p_{\lambda}
\leq
\frac{\delta}{K}.
\]
Since \(p_{\lambda}\) is a valid conservative \(p\)-value under the null hypothesis \(R(\lambda)>\epsilon\), it satisfies
\[
\Pr\left(
p_{\lambda}
\leq
\frac{\delta}{K}
\;\middle|\;
R(\lambda)>\epsilon
\right)
\leq
\frac{\delta}{K}.
\]
Therefore, if the task-failure risk \(R(\lambda)\) exceeds the prescribed target \(\epsilon\), the probability that \(\lambda\) nevertheless passes our certification rule is at most \(\delta/K\). Let
\[
\Lambda^{\mathrm{unsafe}}
=
\left\{
\lambda\in\Lambda:
R(\lambda)>\epsilon
\right\}
\]
denote the set of candidate thresholds that are actually unsafe. The event that at least one threshold whose task-failure risk exceeds \(\epsilon\) is incorrectly certified can then be written as
\[
A
=
\bigcup_{\lambda\in\Lambda^{\mathrm{unsafe}}}
\left\{
\lambda\in
\widehat{\Lambda}_{\mathrm{safe}}
\right\}.
\]
Therefore,
\begin{align*}
\Pr(A)
&=
\Pr\left(
\bigcup_{\lambda\in\Lambda^{\mathrm{unsafe}}}
\left\{
\lambda\in
\widehat{\Lambda}_{\mathrm{safe}}
\right\}
\right)
\\
&\leq
\sum_{\lambda\in\Lambda^{\mathrm{unsafe}}}
\Pr\left(
\lambda\in
\widehat{\Lambda}_{\mathrm{safe}}
\right)
\\
&\leq
\sum_{\lambda\in\Lambda^{\mathrm{unsafe}}}
\frac{\delta}{K}
\\
&=
\left|
\Lambda^{\mathrm{unsafe}}
\right|
\frac{\delta}{K}
\\
&\leq
K\frac{\delta}{K}
=
\delta.
\end{align*}
Equivalently, with probability at least \(1-\delta\), every certified threshold satisfies
\[
R(\lambda)\leq\epsilon.
\]
Therefore,
\[
\Pr\left(
R(\lambda)\leq\epsilon,
\quad
\forall\lambda\in
\widehat{\Lambda}_{\mathrm{safe}}
\right)
\geq
1-\delta.
\]
The deployed threshold
\[
\hat\lambda
\]
is selected from
\[
\widehat{\Lambda}_{\mathrm{safe}}.
\]
Hence, whenever \(A\) does not occur, the deployed threshold must also satisfy
\[
R(\hat\lambda)\leq\epsilon.
\]
It follows that
\[
\Pr\left(
R(\hat\lambda)>\epsilon
\right)
\leq
\delta,
\]
or equivalently,
\[
\Pr\left(
R(\hat\lambda)\leq\epsilon
\right)
\geq
1-\delta.
\]
This completes the proof.
\end{proof}

Theorem~\ref{thm:single-risk} establishes that, for a fixed retrieval buffer, Algorithm~\ref{alg:risk-control} selects the threshold that satisfies the user-specified task-failure requirement with high probability. We next prove Theorem~\ref{thm:continual-risk-main}.

\begin{proof}
Fix an adaptation round \(q\) and let \(\mathcal H_q\) denote the complete deployment history before calibration, including all previous trajectories, \textsc{Act}/\textsc{Ask} decisions, expert corrections, calibrated thresholds, and retrieval buffers. Conditional on \(\mathcal H_q\), the retrieval buffer \(\mathcal D_{\mathrm{ret}}^{(q)}\), the predefined candidate set \(\Lambda\), and the corresponding family of R2-VLA policies \(\{\pi_{\theta,\psi}^{\lambda}:\lambda\in\Lambda\}\) are fixed. By assumption, the calibration episodes are drawn independently from the round-\(q\) deployment distribution conditional on \(\mathcal H_q\). Therefore, Theorem~\ref{thm:single-risk} applies conditionally to round \(q\). In particular, using confidence parameter \(\delta_q\), we have
\[
\Pr
\left(
R_q(\hat\lambda^{(q)})>\epsilon
\mid
\mathcal H_q
\right)
\leq
\delta_q.
\]
This conditional application accounts for the fact that the retrieval buffer may depend on previous adaptation rounds. Once the deployment history is fixed, the updated retrieval buffer and the resulting R2-VLA policy family are also fixed, allowing fresh calibration episodes to provide the same risk-control guarantee as in Theorem~\ref{thm:single-risk}.

Taking the expectation over the deployment history and applying the law of total probability, we obtain
\[
\Pr
\left(
R_q(\hat\lambda^{(q)})>\epsilon
\right)
=
\mathbb E_{\mathcal H_q}
\left[
\Pr
\left(
R_q(\hat\lambda^{(q)})>\epsilon
\mid
\mathcal H_q
\right)
\right]
\leq
\delta_q.
\]
Define the event
\[
E_q
=
\left\{
R_q(\hat\lambda^{(q)})>\epsilon
\right\},
\]
which represents a violation of the prescribed task-failure risk in round \(r\). Applying the union bound across all adaptation rounds yields
\[
\Pr
\left(
\bigcup_{q=0}^{Q}E_q
\right)
\leq
\sum_{q=0}^{Q}\Pr(E_q)
\leq
\sum_{q=0}^{Q}\delta_q.
\]
Taking complements, we obtain
\[
\Pr
\left(
R_q(\hat\lambda^{(q)})\leq\epsilon,
\quad
\forall q=0,\ldots,Q
\right)
\geq
1-\sum_{q=0}^{Q}\delta_q.
\]
Since we allocate the overall confidence budget equally across rounds, setting \(\delta_q=\frac{\delta}{Q+1}\) gives
\[
\sum_{q=0}^{Q}\delta_q=\delta.
\]
Therefore,
\[
\Pr
\left(
R_q(\hat\lambda^{(q)})\leq\epsilon,
\quad
\forall q=0,\ldots,Q
\right)
\geq 1-\delta.
\]
This completes the proof.
\end{proof}

Theorem~\ref{thm:continual-risk-main} establishes that continual growth of the retrieval memory does not invalidate the task-risk control. The retrieval buffer \(\mathcal D_{\mathrm{ret}}^{(q)}\) at round \(q\) may depend arbitrarily on all previous robot trajectories, uncertainty values, \textsc{Ask} decisions, and expert corrections. The key requirement is that, once this updated retrieval buffer is fixed, the threshold is recalibrated using a fresh set of calibration episodes. This distinction is important because adding expert corrections changes the retrieval-augmented policy itself. For the same state \(x_t\), the updated buffer may produce a different retrieved context, action prediction, and uncertainty score. Consequently, a threshold calibrated for \(\mathcal D_{\mathrm{ret}}^{(q)}\) need not remain valid after the memory is updated to \(\mathcal D_{\mathrm{ret}}^{(q+1)}\). Our round-wise recalibration procedure explicitly accounts for this change.

\section{Additional Implementation Details}
\label{app:implement}

\subsection{Benchmark and Evaluation Protocol}
\label{app:benchmark-eval}

We evaluate R2-VLA on LIBERO~\cite{liu2023libero}, a simulated robotic manipulation benchmark based on the Franka Panda arm. Each demonstration contains multi-view visual observations, including front-view and wrist-view images, together with proprioceptive states, natural-language task instructions, and expert actions. We consider all four LIBERO suites: \textsc{Spatial}, \textsc{Object}, \textsc{Goal}, and \textsc{Long}, which evaluate different aspects of manipulation generalization. For each suite, we partition the available demonstrations into disjoint sets for retrieval and policy adaptation, risk calibration, and evaluation. Specifically, demonstration IDs \(10\)--\(29\) are used to construct the retrieval buffer, perform suite-specific RICL adaptation, and train the uncertainty head, corresponding to 20 demonstrations per task and 200 demonstrations per suite. Demonstration IDs \(30\)--\(49\) are reserved for risk calibration, while IDs \(0\)--\(9\) are held out for evaluation and are not used for policy adaptation, uncertainty-head training, or risk calibration.

During deployment, whenever R2-VLA selects \textsc{Ask}, a separately trained expert policy provides the corrective action for the current observation. The expert policy is trained using demonstration IDs \(0\)--\(49\) for each suite and serves exclusively as an action oracle without updating the evaluated VLA during execution. We report task success and ASK rates over 100 evaluation rollouts per task.

For risk calibration, we construct a predefined candidate threshold set \(\Lambda=\{\lambda_1,\ldots,\lambda_K\}\) with \(K=10\). Specifically, we first collect uncertainty scores from five preliminary rollouts using demonstration IDs \(30\)--\(34\). Let \(\mathcal U_{\mathrm{ref}}=\bigcup_{i=1}^{5}\{u_{i,t}\}_{t=1}^{T_i}\) denote the pooled uncertainty scores across these rollouts, where \(T_i\) is the number of decision steps in rollout \(i\). We compute the reference threshold as
\[
\lambda_{\mathrm{ref}}
=
\operatorname{Quantile}_{0.60}
\left(\mathcal U_{\mathrm{ref}}\right).
\]
We then construct \(\Lambda\) by selecting \(K=10\) candidate thresholds evenly spaced over the interval \([\lambda_{\mathrm{ref}}-0.1,\lambda_{\mathrm{ref}}+0.1]\). The resulting candidate set is fixed before applying the risk-control procedure in Algorithm~\ref{alg:risk-control}.

To evaluate continual memory adaptation, we organize deployment into successive adaptation rounds following Algorithm~\ref{alg:continual}. Within each round, the retrieval buffer and calibrated threshold remain fixed, and expert corrections are collected whenever R2-VLA requests assistance. At the end of each round, these corrections are incorporated into the retrieval memory, followed by threshold recalibration using Algorithm~\ref{alg:risk-control}. Specifically, demonstration IDs \(30\)--\(49\) remain designated for calibration, while fresh initial states are independently sampled from the task initialization distribution at each round. We then perform new calibration rollouts using the updated retrieval memory to obtain the calibrated threshold for the next round.

We set the overall confidence parameter to \(\delta=0.05\), corresponding to a \(95\%\) confidence level. For continual memory adaptation with \(Q+1\) rounds, we allocate the confidence budget equally across rounds, using \(\delta_r=\delta/(Q+1)\) for calibration at round \(q\), consistent with Theorem~\ref{thm:continual-risk-main}. All experiments are conducted on a single NVIDIA RTX 4090 GPU with 24\,GB of memory.

\subsection{VLA Backbones}
\label{app:vla-backbone}

We follow the model architecture adopted in RICL~\cite{sridhar2025ricl}, which uses \(\pi_0\)-FAST as the VLA backbone. To evaluate whether R2-VLA generalizes across different VLA architectures and model scales, we additionally consider MiniVLA~\cite{belkhale2024minivla} and OpenVLA~\cite{kim2024openvla}. Across all three backbones, four retrieved demonstrations are provided as in-context examples for each query, i.e., \(M=4\). Because the publicly available backbone from RICL checkpoints are not directly trained for LIBERO experiments, we first adapt each backbone to support retrieval-conditioned execution. 

\paragraph{MiniVLA.}
For MiniVLA, we initialize from the 150,000-step LIBERO checkpoint.\footnote{\url{https://huggingface.co/Stanford-ILIAD/minivla-vq-libero90-prismatic/tree/main/checkpoints}} We first perform 2,000 RICL-priming optimizer updates using 68 LIBERO priming tasks with 20 demonstrations per task. We use rank-32 LoRA, gradient accumulation of 16, AdamW, and a peak learning rate of \(2.5\times10^{-5}\). During priming, the vision encoder and base language-model weights are frozen, while the LoRA parameters, visual projector, and action-token rows remain trainable. Starting from the primed model, we then perform suite-specific adaptation using demonstration IDs \(10\)--\(29\). For each of the four LIBERO suites, we perform 600 additional optimizer updates with a peak learning rate of \(5\times10^{-6}\).

\paragraph{\(\pi_0\)-FAST.}
For \(\pi_0\)-FAST, we initialize from Physical Intelligence's LIBERO-finetuned \(\pi_0\)-FAST checkpoint.\footnote{\url{https://github.com/Physical-Intelligence/openpi}} We directly perform suite-specific RICL adaptation using demonstration IDs \(10\)--\(29\). We use rank-16 LoRA, AdamW, a peak learning rate of \(1\times10^{-5}\), and gradient clipping at 1.0. We use 3,000 updates for \textsc{Object}, 5,000 for \textsc{Spatial}, 13,000 for \textsc{Goal}, and 10,000 for \textsc{Long}.

\paragraph{OpenVLA.}
For OpenVLA, we initialize from the official OpenVLA-7B checkpoint.\footnote{\url{https://huggingface.co/openvla/openvla-7b}} We first perform 1,500 RICL-priming updates using the same 68 priming tasks and 20 demonstrations per task. We use rank-16 LoRA, gradient accumulation of 8, and a peak learning rate of \(5\times10^{-6}\). We subsequently perform 300 additional suite-specific adaptation updates using demonstration IDs \(10\)--\(29\), with rank-16 LoRA, gradient accumulation of 8, and the same peak learning rate of \(5\times10^{-6}\).

\subsection{Uncertainty-Head Implementation}
\label{app:uncertainty-head}

\paragraph{Architecture.}
R2-VLA augments each adapted RICL policy with a lightweight uncertainty head. After RICL adaptation, we freeze the retrieval-conditioned VLA policy and train only the uncertainty head. This separation ensures that uncertainty-head training does not modify the underlying action policy. We extract the final-layer hidden representation at the action-start position, corresponding to the last conditioning-token representation after the model's final normalization and immediately before the vocabulary projection. A lightweight MLP maps this representation to a nonnegative scalar \(u\), which estimates the action prediction error of the retrieval-conditioned VLA. We train the uncertainty head to minimize
\[
\mathcal{L}_U
=
\frac{1}{\mathbb{E}[T]}
\mathbb{E}\left[
\sum_{t=1}^{T}
\mathrm{SmoothL1}\bigl(
\log(\hat u_t+\eta),\log(u_t+\eta)
\bigr)
\right],
\]
where \(\hat u_t\) is the uncertainty-head output, \(u_t\) is the normalized action-chunk RMSE, and \(\eta=10^{-4}\) is used for numerical stability.

\paragraph{Training Data.}
For all backbones and LIBERO suites, uncertainty-head training uses demonstration IDs \(10\)--\(29\), i.e., the same demonstrations used for retrieval and suite-specific policy adaptation. Training exclusively on clean retrieval contexts provides limited supervision for states in which retrieval is misleading. We therefore augment the training data with deliberately corrupted retrieval contexts. These corruptions are designed to expose the uncertainty head to situations in which the retrieved examples may appear plausible but provide incorrect or uninformative guidance for the current action. We consider four corruption types:

\begin{enumerate}
    \item \textbf{Wrong-task retrieval.}
    The retrieved example is sampled from a different task, preferentially from a task whose instruction shares words with the current task but whose scene or intended behavior differs.

    \item \textbf{Correct task, wrong phase.}
    The retrieval is taken from the correct task and demonstration, but the retrieved timestep is shifted forward. The retrieved action therefore corresponds to a different stage of task execution.

    \item \textbf{Visually similar, wrong goal.}
    The retrieved example is selected from another task in the same scene, producing a visually plausible observation while corresponding to a different task goal and intended behavior.

    \item \textbf{Random retrieval.}
    The retrieved example is sampled randomly from a different task, providing context that is largely unrelated to the current decision.
\end{enumerate}

Among corrupted training examples, the four corruption types are sampled according to the proportions in Table~\ref{tab:composition-corrupt}. The fraction of uncertainty-head training examples that use corrupted retrieval contexts is backbone dependent. We use 30\% corrupted contexts for MiniVLA, 50\% for \(\pi_0\)-FAST, and 30\% for OpenVLA. the remaining examples use the original clean retrieval context.

\begin{table}[t]
\centering
\small
\caption{Composition of corrupted retrieval examples used to train the uncertainty head.}
\label{tab:composition-corrupt}
\begin{tabular}{lc}
\toprule
\textbf{Corruption Type} & \textbf{Share of Corrupted} \\
\midrule
Wrong-task retrieval & 25\% \\
Correct task, wrong phase & 25\% \\
Visually similar, wrong goal & 35\% \\
Random/uninformative retrieval & 15\% \\
\bottomrule
\end{tabular}
\end{table}

\paragraph{Training Hyperparameters.}
Table~\ref{tab:uncertainty_hparams} summarizes the optimization hyperparameters for uncertainty head training.

\begin{table}[t]
\centering
\small
\caption{Hyperparameters for uncertainty-head training.}
\label{tab:uncertainty_hparams}
%\resizebox{\linewidth}{!}{%
\begin{tabular}{lccc}
\toprule
\textbf{Hyperparameter}
& \textbf{MiniVLA}
& \textbf{\(\pi_0\)-FAST}
& \textbf{OpenVLA} \\
\midrule
Optimizer
& AdamW
& AdamW
& AdamW \\
Learning Rate
& \(1\times10^{-4}\)
& \(3\times10^{-4}\)
& \(1\times10^{-4}\) \\
Weight Decay
& \(1\times10^{-4}\)
& \(1\times10^{-4}\)
& \(1\times10^{-4}\) \\
Grad. Clip
& 5.0
& 5.0
& 5.0 \\
Updates
& 3,000
& 2,500
& 1,000 \\
\bottomrule
\end{tabular}
%}
\end{table}

\section{Additional Experimental Results}
\label{app:results}

\begin{table*}[t]
\centering
\caption{
Comparison of expert assistance strategies on LIBERO using MiniVLA and OpenVLA backbones. SR denotes success rate.
}
\label{tab:intervention-baselines-additional}
\scriptsize
\setlength{\tabcolsep}{2.5pt}
\renewcommand{\arraystretch}{1.08}
\resizebox{\textwidth}{!}{%
\begin{tabular}{ll ccc ccc ccc ccc cc}
\toprule
\multirow{2}{*}{Backbone}
& \multirow{2}{*}{Expert Assistance Strategy}
& \multicolumn{3}{c}{Spatial}
& \multicolumn{3}{c}{Object}
& \multicolumn{3}{c}{Goal}
& \multicolumn{3}{c}{Long}
& \multirow{2}{*}{Avg. SR $\uparrow$}
& \multirow{2}{*}{Avg. ASK $\downarrow$}
\\
\cmidrule(lr){3-5}
\cmidrule(lr){6-8}
\cmidrule(lr){9-11}
\cmidrule(lr){12-14}
&
& SR $\uparrow$ & Target SR & ASK $\downarrow$
& SR $\uparrow$ & Target SR & ASK $\downarrow$
& SR $\uparrow$ & Target SR & ASK $\downarrow$
& SR $\uparrow$ & Target SR & ASK $\downarrow$
& &
\\
\midrule

MiniVLA
& RICL (no help)
& 58.9\% & -- & 0.0\%
& 59.1\% & -- & 0.0\%
& 60.6\% & -- & 0.0\%
& 28.5\% & -- & 0.0\%
& 51.8\% & 0.0\%
\\

& Random ASK
& 61.3\% & 65.0\% & 17.0\%
& 63.3\% & 65.0\% & 16.9\%
& 64.9\% & 65.0\% & 14.4\%
& 30.1\% & 35.0\% & 34.9\%
& 54.9\% & 20.8\%
\\

& DINO Distance + Risk Control
& 66.1\% & 65.0\% & 30.2\%
& 67.7\% & 65.0\% & 29.5\%
& 65.4\% & 65.0\% & 28.8\%
& 34.6\% & 35.0\% & 47.6\%
& 58.5\% & 34.0\%
\\

& Action Entropy + Risk Control
& 65.2\% & 65.0\% & 31.4\%
& 68.9\% & 65.0\% & 28.6\%
& 66.4\% & 65.0\% & 29.1\%
& 34.7\% & 35.0\% & 46.9\%
& 58.8\% & 34.0\%
\\

\rowcolor{lightblue!100}
& \textbf{Learned Uncertainty + Risk Control (ours)}
& \textbf{67.1\%} & 65.0\% & 17.0\%
& \textbf{77.4\%} & 65.0\% & 16.9\%
& \textbf{71.0\%} & 65.0\% & 14.4\%
& \textbf{35.7\%} & 35.0\% & 34.9\%
& \textbf{62.8\%} & \textbf{20.8\%}
\\

\midrule

OpenVLA
& RICL (no help)
& 67.3\% & -- & 0.0\%
& 65.8\% & -- & 0.0\%
& 64.7\% & -- & 0.0\%
& 31.6\% & -- & 0.0\%
& 57.4\% & 0.0\%
\\

& Random ASK
& 69.4\% & 75.0\% & 31.8\%
& 68.7\% & 75.0\% & 30.1\%
& 67.9\% & 75.0\% & 26.8\%
& 35.2\% & 40.0\% & 44.6\%
& 60.3\% & 33.3\%
\\

& DINO Distance + Risk Control
& 77.2\% & 75.0\% & 48.9\%
& 75.1\% & 75.0\% & 49.7\%
& 76.4\% & 75.0\% & 40.2\%
& 38.6\% & 40.0\% & 66.9\%
& 66.8\% & 51.4\%
\\

& Action Entropy + Risk Control
& 77.4\% & 75.0\% & 49.1\%
& 75.8\% & 75.0\% & 48.5\%
& 76.2\% & 75.0\% & 41.2\%
& 39.7\% & 40.0\% & 67.4\%
& 67.3\% & 51.6\%
\\

\rowcolor{lightblue!100}
& \textbf{Learned Uncertainty + Risk Control (ours)}
& \textbf{79.3\%} & 75.0\% & 31.8\%
& \textbf{76.5\%} & 75.0\% & 30.1\%
& \textbf{78.4\%} & 75.0\% & 26.8\%
& \textbf{40.2\%} & 40.0\% & 44.6\%
& \textbf{68.6\%} & \textbf{33.3\%}
\\

\bottomrule
\end{tabular}
}
\end{table*}

\begin{figure*}[t]
    \centering
    \begin{subfigure}[t]{0.32\textwidth}
        \centering
        \includegraphics[width=\linewidth]{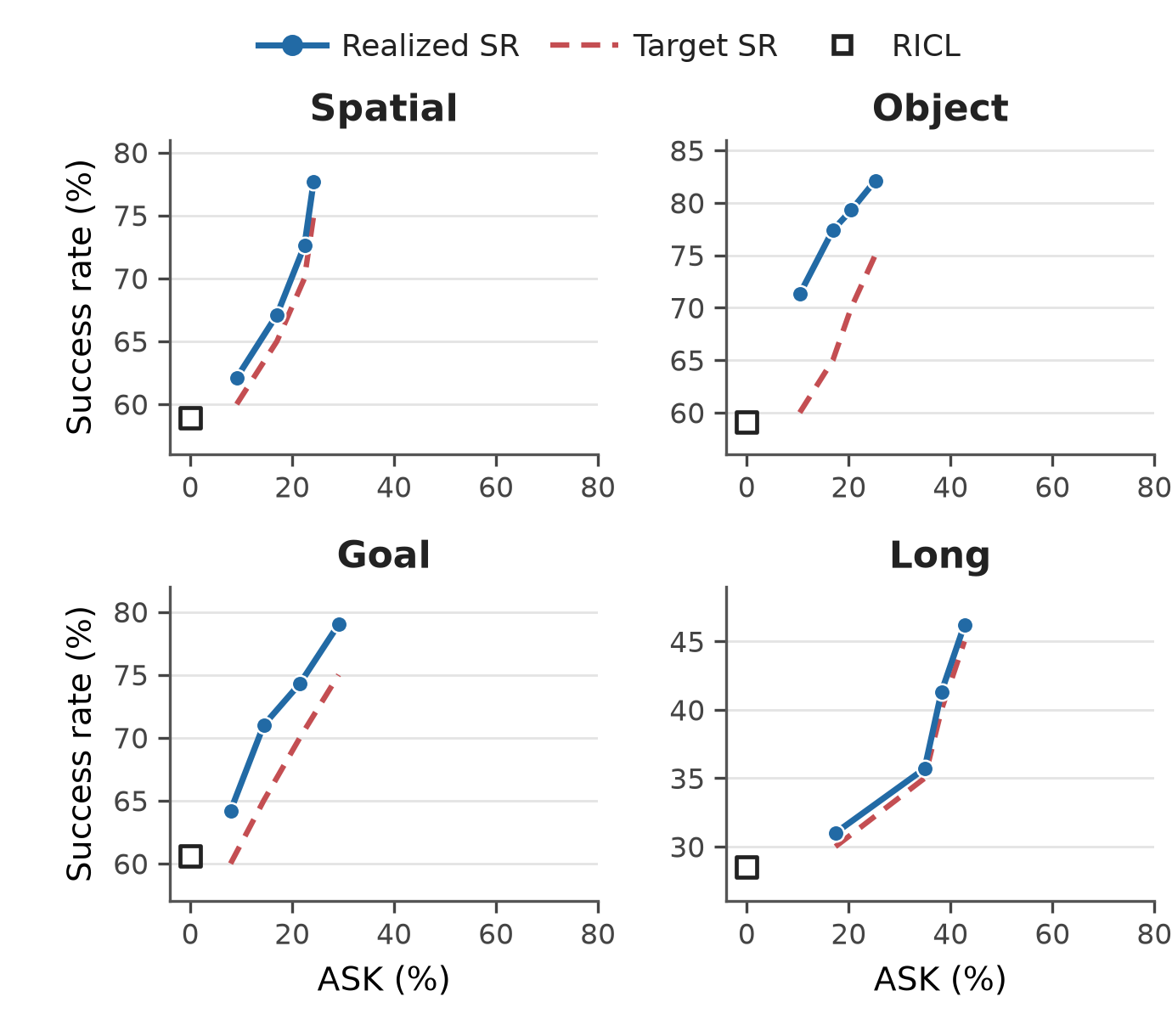}
        \caption{MiniVLA}
        \label{fig:risk-control-minivla}
    \end{subfigure}
    \hfill
    \begin{subfigure}[t]{0.32\textwidth}
        \centering
        \includegraphics[width=\linewidth]{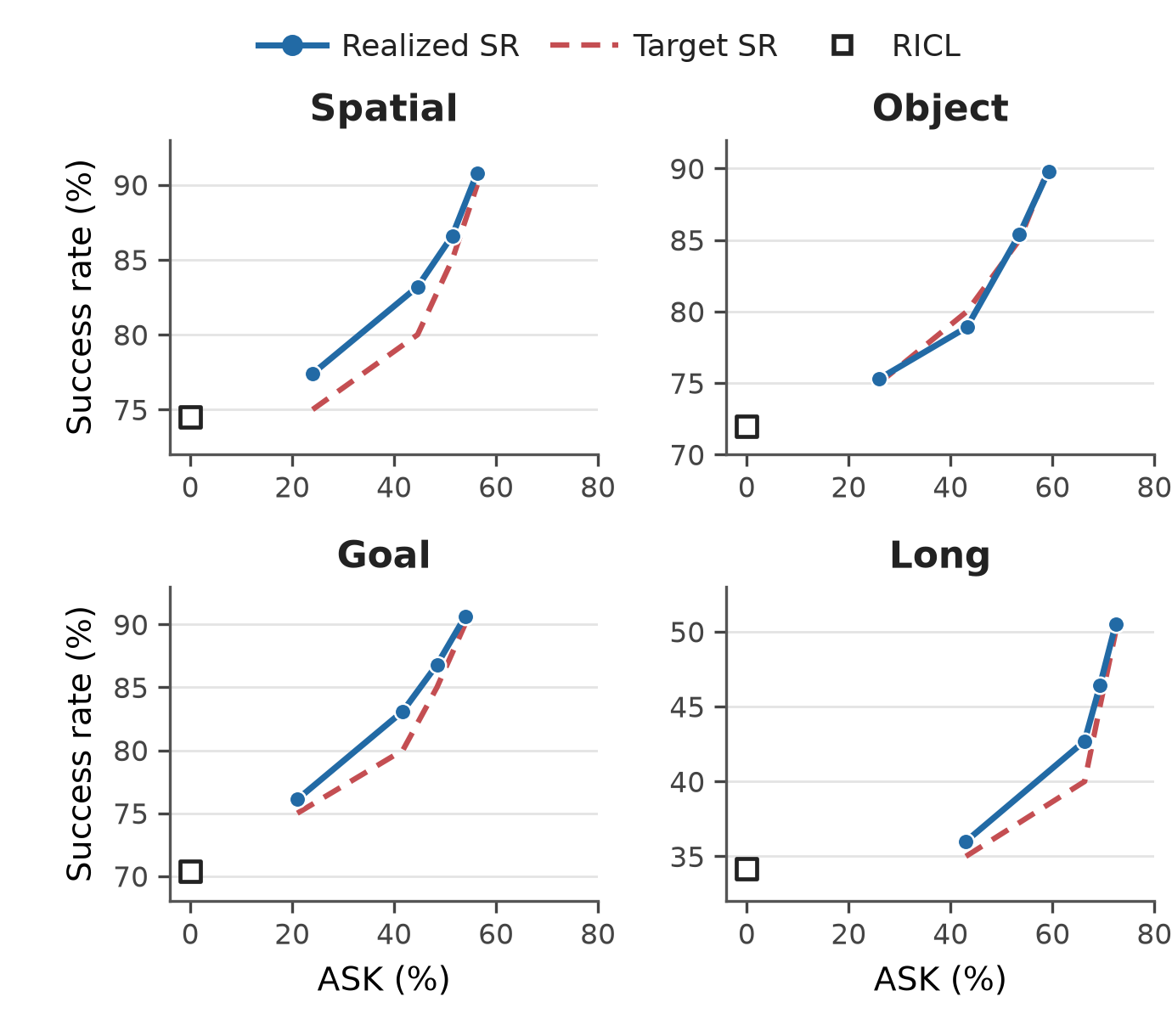}
        \caption{$\pi_0$-FAST}
        \label{fig:risk-control}
    \end{subfigure}
    \hfill
    \begin{subfigure}[t]{0.32\textwidth}
        \centering
        \includegraphics[width=\linewidth]{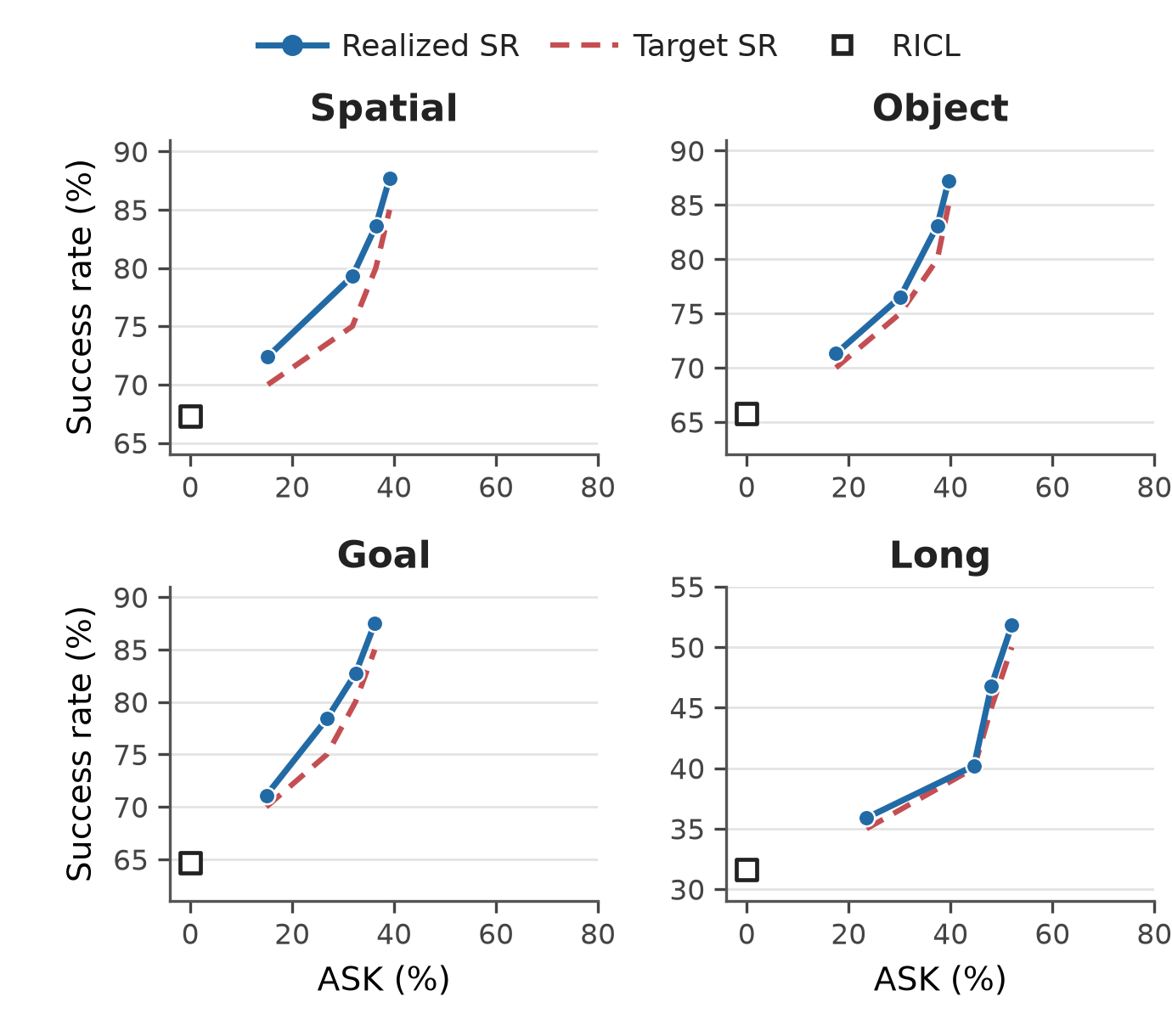}
        \caption{OpenVLA}
        \label{fig:risk-control-openvla}
    \end{subfigure}

    \caption{
    Effect of different target risk levels on R2-VLA across three VLA backbones and four LIBERO suites. As the target success rate increases, R2-VLA generally requests more expert assistance to achieve higher task success, demonstrating the trade-off between task-level success and the frequency of expert assistance.
    }
    \label{fig:risk-control-all}
    \vspace{-2ex}
\end{figure*}

\begin{table*}[t]
\centering
\caption{
Ablation of retrieval corruption for uncertainty-head training across VLA backbones. SR denotes success rate.
}
\label{tab:ablation-corruption-all}
\scriptsize
\setlength{\tabcolsep}{2.5pt}
\renewcommand{\arraystretch}{1.08}
\resizebox{\textwidth}{!}{%
\begin{tabular}{ll ccc ccc ccc ccc cc}
\toprule
\multirow{2}{*}{Backbone}
& \multirow{2}{*}{Variant}
& \multicolumn{3}{c}{Spatial}
& \multicolumn{3}{c}{Object}
& \multicolumn{3}{c}{Goal}
& \multicolumn{3}{c}{Long}
& \multirow{2}{*}{Avg. SR $\uparrow$}
& \multirow{2}{*}{Avg. ASK $\downarrow$}
\\
\cmidrule(lr){3-5}
\cmidrule(lr){6-8}
\cmidrule(lr){9-11}
\cmidrule(lr){12-14}
&
& SR $\uparrow$ & Target SR & ASK $\downarrow$
& SR $\uparrow$ & Target SR & ASK $\downarrow$
& SR $\uparrow$ & Target SR & ASK $\downarrow$
& SR $\uparrow$ & Target SR & ASK $\downarrow$
& &
\\
\midrule

MiniVLA
& w/o Retrieval Corruption
& 66.4\% & 65.0\% & 29.1\%
& 69.3\% & 65.0\% & 28.3\%
& 67.1\% & 65.0\% & 29.4\%
& 35.2\% & 35.0\% & 48.6\%
& 59.5\% & 33.9\%
\\

\rowcolor{lightblue!100}
& \textbf{R2-VLA}
& \textbf{67.1\%} & 65.0\% & 17.0\%
& \textbf{77.4\%} & 65.0\% & 16.9\%
& \textbf{71.0\%} & 65.0\% & 14.4\%
& \textbf{35.7\%} & 35.0\% & 34.9\%
& \textbf{62.8\%} & \textbf{20.8\%}
\\

\midrule

\(\pi_0\)-FAST
& w/o Retrieval Corruption
& 82.3\% & 80.0\% & 55.1\%
& 76.7\% & 80.0\% & 54.7\%
& 81.8\% & 80.0\% & 57.6\%
& 41.2\% & 40.0\% & 79.3\%
& 70.5\% & 61.7\%
\\

\rowcolor{lightblue!100}
& \textbf{R2-VLA}
& \textbf{83.2\%} & 80.0\% & 44.6\%
& \textbf{78.9\%} & 80.0\% & 43.2\%
& \textbf{83.1\%} & 80.0\% & 41.7\%
& \textbf{42.7\%} & 40.0\% & 66.3\%
& \textbf{72.0\%} & \textbf{49.0\%}
\\

\midrule

OpenVLA
& w/o Retrieval Corruption
& 76.7\% & 75.0\% & 46.4\%
& 75.3\% & 75.0\% & 45.3\%
& 76.5\% & 75.0\% & 41.2\%
& 40.3\% & 40.0\% & 56.8\%
& 67.2\% & 47.4\%
\\

\rowcolor{lightblue!100}
& \textbf{R2-VLA}
& \textbf{79.3\%} & 75.0\% & 31.8\%
& \textbf{76.5\%} & 75.0\% & 30.1\%
& \textbf{78.4\%} & 75.0\% & 26.8\%
& \textbf{40.2\%} & 40.0\% & 44.6\%
& \textbf{68.6\%} & \textbf{33.3\%}
\\

\bottomrule
\end{tabular}
}
\end{table*}

\begin{table*}[t]
\centering
\caption{
Effect of calibration-set size on R2-VLA using the $\pi_0$-FAST backbone across four LIBERO suites. SR denotes task success rate, and ASK denotes expert assistance frequency. The highlighted row corresponds to the default calibration-set size used in the main experiments.
}
\label{tab:ablation-calibration-size}
\scriptsize
\setlength{\tabcolsep}{3pt}
\renewcommand{\arraystretch}{1.08}
\resizebox{\textwidth}{!}{%
\begin{tabular}{c ccc ccc ccc ccc cc}
\toprule
\multirow{2}{*}{$N_{\mathrm{cal}}$}
& \multicolumn{3}{c}{Spatial}
& \multicolumn{3}{c}{Object}
& \multicolumn{3}{c}{Goal}
& \multicolumn{3}{c}{Long}
& \multirow{2}{*}{Avg. SR $\uparrow$}
& \multirow{2}{*}{Avg. ASK $\downarrow$}
\\
\cmidrule(lr){2-4}
\cmidrule(lr){5-7}
\cmidrule(lr){8-10}
\cmidrule(lr){11-13}
& SR $\uparrow$ & Target SR & ASK $\downarrow$
& SR $\uparrow$ & Target SR & ASK $\downarrow$
& SR $\uparrow$ & Target SR & ASK $\downarrow$
& SR $\uparrow$ & Target SR & ASK $\downarrow$
& &
\\
\midrule

50
& 84.0\% & 80.0\% & 51.2\%
& 79.0\% & 80.0\% & 49.3\%
& 83.8\% & 80.0\% & 49.6\%
& 42.2\% & 40.0\% & 70.1\%
& 72.3\% & 55.1\%
\\

100
& 83.6\% & 80.0\% & 50.4\%
& 78.9\% & 80.0\% & 48.6\%
& 83.4\% & 80.0\% & 47.9\%
& 42.9\% & 40.0\% & 69.8\%
& 72.2\% & 54.2\%
\\

\rowcolor{lightblue!100}
\textbf{200 (Default)}
& 83.2\% & 80.0\% & 44.6\%
& 78.9\% & 80.0\% & 43.2\%
& 83.1\% & 80.0\% & 41.7\%
& 42.7\% & 40.0\% & 66.3\%
& \textbf{72.0\%} & \textbf{49.0\%}
\\

400
& 83.7\% & 80.0\% & 40.8\%
& 79.3\% & 80.0\% & 40.5\%
& 83.9\% & 80.0\% & 39.6\%
& 43.0\% & 40.0\% & 62.5\%
& 72.5\% & 45.9\%
\\

\bottomrule
\end{tabular}
}
\end{table*}

\subsection{Expert Assistance Strategies Baselines on MiniVLA and OpenVLA.} 
\label{app:intervention-baseline}

We evaluate the effectiveness of different expert assistance strategies on MiniVLA and OpenVLA, with the complete results reported in Table~\ref{tab:intervention-baselines-additional}. Consistent with the findings for \(\pi_0\)-FAST in the main paper, R2-VLA achieves the highest average task success rate while requiring substantially fewer expert assistance than alternative uncertainty-based strategies. Specifically, on MiniVLA, R2-VLA achieves an average success rate of 62.8\% with an ASK rate of only 20.8\%, outperforming DINO Distance (58.5\% success with 34.0\% ASK) and Action Entropy (58.8\% success with 34.0\% ASK). Similarly, on OpenVLA, R2-VLA achieves 68.6\% success with 33.3\% ASK, compared with 66.8\% success and 51.4\% ASK for DINO Distance, and 67.3\% success and 51.6\% ASK for Action Entropy. These results demonstrate that the learned uncertainty identifies more informative decision points, allowing R2-VLA to achieve higher task success with less expert assistance.

Furthermore, when compared with Random ASK under matched ASK budgets, R2-VLA improves average success by 7.9 percentage points on MiniVLA (62.8\% versus 54.9\%) and 8.3 percentage points on OpenVLA (68.6\% versus 60.3\%). This indicates that the improvements cannot be attributed solely to the additional expert actions, but rather to the ability of the learned uncertainty to identify when autonomous execution is uncertain. Together with the \(\pi_0\)-FAST results in the main paper, these findings demonstrate that R2-VLA consistently improves task success while reducing the frequency of expert assistance across different VLA architectures and model scales.

\subsection{Ablation Studies}
\label{app:ablation}

\paragraph{Effect of Target Risk Level.}
We investigate how different user-specified risk levels \(\epsilon\) affect the performance of R2-VLA. Specifically, we vary the target success rate \(1-\epsilon\) and independently calibrate the corresponding threshold \(\hat\lambda\). Figure~\ref{fig:risk-control-all} illustrates the resulting trade-offs between task success and ASK rate for \(\pi_0\)-FAST, MiniVLA, and OpenVLA, respectively, across all four LIBERO suites. As the target success rate increases, R2-VLA generally requests more expert assistance, leading to improved task success. Importantly, the observed success rates closely track or exceed the prescribed targets across different risk levels, demonstrating that R2-VLA can adapt its ACT/ASK decisions to varying task success requirements. The results also reveal differences across task suites: LIBERO-\textsc{Long} generally requires more frequent expert assistance to achieve a given success rate, reflecting the greater difficulty of long-horizon manipulation tasks. These trends are consistent across all three VLA backbones, demonstrating that R2-VLA enables flexible control over task-level success rate by adjusting the calibrated threshold according to the desired risk level, without relying on a manually specified uncertainty threshold.

\paragraph{Retrieval Corruption.}

We evaluate the contribution of corrupted retrieval contexts during uncertainty-head training by comparing R2-VLA against a variant trained exclusively with clean retrieval demonstrations. As shown in Table~\ref{tab:ablation-corruption-all}, incorporating retrieval corruption consistently improves average task success while substantially reducing ASK rate across all three VLA backbones. Specifically, for MiniVLA, the average success rate increases from 59.5\% to 62.8\%, while the ASK rate decreases from 33.9\% to 20.8\%. Similar improvements are observed for \(\pi_0\)-FAST, where the success rate increases from 70.5\% to 72.0\% and the ASK rate decreases from 61.7\% to 49.0\%, and for OpenVLA, where the success rate increases from 67.2\% to 68.6\% while the ASK rate decreases from 47.4\% to 33.3\%. These results suggest that exposing the uncertainty head to misleading retrieval contexts improves its ability to distinguish certain from uncertain retrieval-guided actions. Importantly, the reduction in ASK rate is accompanied by improved task success, indicating that retrieval corruption does not merely encourage more conservative uncertainty estimates but instead enables more selective expert assistance. Overall, these findings support the importance of explicitly incorporating retrieval failure modes into uncertainty-head training to improve both task success rate and expert assitance frequency.

\paragraph{Effect of Calibration Set Size.}
We investigate the sensitivity of R2-VLA to the number of calibration episodes \(N_{\mathrm{cal}}\). Specifically, we vary \(N_{\mathrm{cal}}\in\{50,100,200,400\}\) while keeping the other parameter fixed. As shown in Table~\ref{tab:ablation-calibration-size}, increasing the calibration-set size consistently reduces the frequency of expert assistance while maintaining relatively stable task success across all four LIBERO suites. In particular, increasing \(N_{\mathrm{cal}}\) from \(50\) to \(400\) reduces the average ASK rate from \(55.1\%\) to \(45.9\%\), while the average success rate remains around \(72\%\). This suggests that additional calibration data improves the ability of the risk-control procedure to identify thresholds that permit more autonomous execution without substantially compromising task success. Notably, the default setting of \(N_{\mathrm{cal}}=200\), used in our main experiments, already achieves a favorable trade-off between task success and expert assistance, with an average success rate of \(72.0\%\) and an ASK rate of \(49.0\%\). These findings demonstrate that R2-VLA can maintain relatively stable task performance across different calibration-set sizes while reducing unnecessary expert assistance as more calibration data becomes available.

\end{document}